\def\year{2021}\relax
\documentclass[letterpaper]{article} 
\usepackage{aaai21}  
\usepackage{times}  
\usepackage{helvet} 
\usepackage{courier}  
\usepackage[hyphens]{url}  
\usepackage{graphicx} 
\usepackage{natbib}  
\usepackage{caption} 
\usepackage{lipsum}
\usepackage{array}
\usepackage{graphicx}
\usepackage{dcolumn}
\usepackage{multirow}
\usepackage{paralist} 
\usepackage{subcaption}
\usepackage{makecell,xcolor,amsmath}
\usepackage{colortbl}
\usepackage{forest} 
\usepackage[ruled,lined,linesnumbered]{algorithm2e}

\SetCommentSty{mycommfont}
\SetKwInOut{AlgInput}{input}
\SetKwInOut{AlgOutput}{output}
\SetKwComment{AlgComment}{// }{}
\usepackage{hyperref}
\usepackage{xcolor}
\usepackage{float}
\usepackage{ctable}
\usepackage[section]{placeins}
\hypersetup{
	colorlinks=false,
	linkcolor={red!20!black},
	citecolor={green!20!black},
	urlcolor={blue!20!black}
}
\newcommand{\hideh}[1]{}

\AtBeginDocument{%
  \providecommand\BibTeX{{%
    \normalfont B\kern-0.5em{\scshape i\kern-0.25em b}\kern-0.8em\TeX}}}

\title{Does Fair Ranking Improve Minority Outcomes?\\
Understanding the Interplay of Human and Algorithmic Biases in Online Hiring}
\author {
    Tom S\"uhr,\textsuperscript{\rm 1}
    Sophie Hilgard, \textsuperscript{\rm 2}
    Himabindu Lakkaraju \textsuperscript{\rm 2} \\
}
\affiliations {
    \textsuperscript{\rm 1} Technische Universit\"at Berlin \\
    \textsuperscript{\rm 2} Harvard University \\
    tom.suehr@googlemail.com, ash798@g.harvard.edu, hlakkaraju@hbs.edu
}

\begin{document}
\maketitle
\newcount\Comments
\Comments = 1
\newcommand{\kibitz}[2]{\ifnum\Comments=1{\color{#1}{#2}}\fi}
\newcommand{\tom}[1]{\kibitz{blue}{[TOM: #1]}}
\newcommand{\hima}[1]{\kibitz{red}{[HIMA: #1]}}
\newcommand{\sophie}[1]{\kibitz{purple}{[SOPHIE: #1]}}
\newcommand{\todo}[1]{\kibitz{red}{[TODO: #1]}}
\newcommand{\toref}{\kibitz{red}{[REF]}}

\newcommand{\rabbit}[0]{{\sc RabbitRanking}\xspace}
\newcommand{\rabbitswapped}[0]{{\sc RabbitRanking(F$\leftrightarrow$M)}\xspace}
\newcommand{\random}[0]{{\sc RandomRanking}\xspace}
\newcommand{\randomswapped}[0]{{\sc RandomRanking(F$\leftrightarrow$M)}\xspace}
\newcommand{\fair}[0]{{\sc FairDet-Greedy}\xspace}
\newcommand{\fairswapped}[0]{{\sc FairDet-Greedy(F$\leftrightarrow$M)}\xspace}

\newcommand{\candidate}{candidate\xspace}
\newcommand{\employer}{employer\xspace}
\newcommand{\jobcontext}{job context\xspace}
\newcommand{\underrepresented}{underrepresented\xspace}
\newcommand{\Underrepresented}{Underrepresented\xspace}
\newcommand{\underrepresentation}{underrepresentation\xspace}

\newcommand{\candidates}{candidates\xspace}
\newcommand{\employers}{employers\xspace}
\newcommand{\jobcontexts}{job contexts\xspace}

\definecolor{-5-10}{RGB}{214,96,77}
\definecolor{-5}{RGB}{244,165,130}
\definecolor{<5}{RGB}{209,229,240}
\definecolor{5-10}{RGB}{146,197,222}
\definecolor{10-15}{RGB}{67,147,195}
\definecolor{>15}{RGB}{33,102,172}
\definecolor{0}{RGB}{255,255,255}
\definecolor{ba}{RGB}{230, 230, 230}
\definecolor{he}{RGB}{181, 181, 181}

\begin{abstract}
Ranking algorithms are being widely employed in various online hiring platforms including LinkedIn, TaskRabbit, and Fiverr.
Prior research has demonstrated that ranking algorithms employed by these platforms are prone to a variety of undesirable biases, leading to the proposal of fair ranking algorithms (e.g., \emph{Det-Greedy}) which increase exposure of \underrepresented \candidates.
However, there is little to no work that explores whether 
fair ranking algorithms actually improve real world 
outcomes (e.g., hiring decisions) for \underrepresented groups. Furthermore, there is no clear understanding as to how other factors (e.g., \jobcontext, inherent biases of the \employers) may impact the efficacy of fair ranking in practice.

In this work, we 
analyze various sources of gender biases in online hiring platforms, including the 
\jobcontext and inherent biases of \employers and establish how these factors interact with ranking algorithms to affect hiring decisions. To the best of our knowledge, this work makes the first attempt at studying the interplay between the aforementioned factors in the context of online hiring. 
We carry out a large-scale user study simulating online hiring scenarios with data from TaskRabbit, a popular online freelancing site.
Our results demonstrate that while fair ranking algorithms generally improve the selection rates of \underrepresented minorities, their effectiveness relies heavily on the \jobcontexts and \candidate profiles. 
\end{abstract}

\section{Introduction}\label{sec:introduction}
Over the past decade, there has been a dramatic increase in usage of online hiring platforms and marketplaces such as LinkedIn, TaskRabbit, and Fiverr. 
These platforms are powered by automated tools and algorithms that determine how job seekers are presented to potential \employers, e.g. by filtering and ranking available \candidates. 
Since such platforms impact job seekers' livelihood, it is critical to ensure that the underlying algorithms are not adversely affecting \underrepresented groups.
However, recent research has demonstrated that ranking algorithms employed by various online platforms tend to amplify undesirable biases~\cite{hannak2017bias}. 

\hideh{
\hima{Motivation}
-- Rankings most common way to present recommendations on platforms (Web search, hiring, music, dating, social media, ...)
-- In crowd sourcing and hiring, peoples livelihood depend on those recommendations
-- Concerns about biases and unfairness on those platforms \cite{hannak2017bias}
-- Feedback loops in recommendation systems can amplify bias \cite{jiang2019degenerate}.}

Emerging work in algorithmic fairness tackles the aforementioned challenges by proposing \emph{fair ranking algorithms}, which adjust relevance-only rankings to redistribute user attention across groups or individuals in an equitable fashion~\citep{singh2019policy, morik2020controlling, singh2018fairness, zehlike2020reducing}. Different notions of fairness have been proposed; For example, ~\citet{zehlike2017fa} optimize for a \emph{group fairness} criterion by proposing a post-processing approach which 
ensures that the representation of the \underrepresented group does not fall below a minimum threshold at any point in the ranked list. On the other hand,~\citet{biega2018equity} formalize an \emph{individual equity-of-attention} notion of fairness, proposing 
to fairly divide attention between equally relevant candidates. 

While theoretically promising, 
these approaches have been evaluated only through the simulation of web search-based click models \cite{chuklin2015click}, 
and have not been 
validated in real world settings (e.g., hiring decisions in online portals). 
Furthermore, there is little to no research that systematically explores how other factors (e.g., inherent biases of \employers) may interact with fair ranking algorithms and impact real world hiring decisions.
\hideh{
\hima{Prior Work and Gaps}
-- Algorithmic fairness community tries to mitigate those concerns through redistribution of attention/exposure
-- Many approaches tackle different fairness definitions for statistical and deterministic \sophie{also stochastic?} rankings \cite{biega2018equity}, \cite{joachims2017accurately}, \cite{zehlike2017fa}, \cite{celis2020interventions}. 
-- But for users/workers, the outcome is important
-- We don't know much about effects of changing an algorithm on decision outcome in the context of rankings and hiring
-- Geyik et al. \cite{geyik2019fairness} reported that the deployment of a fair ranking algorithm, search results on LinkedIn were more representative and they observed no significant
-- Peng et al. \cite{peng2019you} investigated human decision making in hiring decisions and found that over representation of candidates in different contexts can work, however not in a ranking scenario.
-- Jahanbakhsh et al. \cite{jahanbakhsh2020experimental} showed that users are biased in giving platform workers star ratings

\hima{Why is it Challenging?}
}

In this work, we address the aforementioned gaps in existing literature by studying 
how various sources of gender biases in online hiring platforms such as the \jobcontext and inherent biases of \employers interact with each other and with ranking algorithms to affect hiring decisions. By studying this interplay, we provide answers to some critical and fundamental questions which have not been systematically explored in existing literature: 1) Do \employers exhibit gender bias uniformly across all \jobcontexts and \candidates? Or do certain kinds of \jobcontexts promote gender biases more than others? 2) What kinds of ranking algorithms are effective in mitigating gender biases in hiring decisions? 3) Can fair ranking algorithms lead to disparate outcomes for different underrepresented groups? 
To the best of our knowledge, this work makes the first attempt at 
studying the interactions between various factors in online hiring such as ranking algorithms, \jobcontexts, and \candidate profiles, and analyzing how they collectively impact hiring decisions. 

To answer the aforementioned questions, we carried out a large-scale user study to simulate hiring scenarios on TaskRabbit, a popular online freelancing platform. We recruited 1,079 participants on Amazon Mechanical Turk and leveraged real world data from TaskRabbit to carry out this study. 
Each participant served as a proxy \employer and was required to select among ranked
\candidates to help them with three different tasks in our controlled hiring platform,
in which we systematically vary ranking algorithms and \candidate features.
We 
use the responses collected from 
this study to carry out our analysis and answer critical questions about the propagation of gender biases in online hiring. 

Our analysis revealed 
that fair ranking algorithms can be helpful in increasing the number of \underrepresented \candidates selected. 
However, their effectiveness is dampened in those \jobcontexts where \employers have a persistent gender preference. We find that fair ranking is more effective when \underrepresented \candidate profiles (features) are similar to those of the majority class. 
Further, we find evidence that fair ranking is ineffective at increasing minority representation when \employer choices already satisfy equal selection rates. Interestingly, we find that some \employers knowingly apply their own notions of fairness not only within tasks (e.g. selecting from the bottom to ``give inexperienced candidates a chance") but also across tasks by knowingly discriminating against a group of \candidates in one \jobcontext and ``making up for it" in another \jobcontext. 
While we carry out our studies with data from TaskRabbit, we believe that our findings generalize to other online hiring portals (e.g., Upwork, Thumbtack, Fiverr) which share similar characteristics. 
\section{Related Work}\label{sec:related-work}

Our work spans multiple topics under the broad umbrella of fairness and bias detection. More specifically, our work lies at the intersection of: 1) empirical evidence of gender bias in online portals, 2) fair ranking algorithms and their effectiveness, and  3) user-algorithm interaction. We discuss related work on each of these topics in detail below. \\
\\
\textbf{Empirical Evidence of Gender Bias}
The existence of gender bias in hiring and evaluation settings has been well documented both in online settings and in the real world. 
For instance,
\citet{hannak2017bias} empirically established the presence of gender and racial biases in reviews and ratings on online marketplaces such as TaskRabbit and Fiverr. They found that female \candidates receive fewer reviews on TaskRabbit compared to their male counterparts with equivalent experience. 
They also found evidence that Black \candidates receive worse ratings on TaskRabbit, and both worse ratings and fewer reviews on Fiverr. 
\citet{nieva1980sex} studied gender biases in evaluations and found strong evidence for pro-male bias. 

More recently, 
\citet{jahanbakhsh2020experimental} investigated the interaction of gender and performance on worker ratings in a simulated teamwork task on Amazon Mechanical Turk. They found that when male and female coworkers were equally low performing, the female worker received worse evaluations. 
Furthermore, \citet{peng2019you} 
found that increasing the representation of \underrepresented \candidates can sometimes correct for biases caused by a skewed \candidate distribution, but human biases in certain \jobcontexts persist even after increasing representation of the \underrepresented group. However, they investigated the effects of increased representation in a non-ranking scenario.
Furthermore, additional work has not only investigated gender bias in contexts other than hiring \cite{may2019gender, shakespeare2020exploring, ekstrand2018exploring, kay2015unequal} but also shed light on other kinds of biases in the context of hiring \cite{thebault2015avoiding, bertrand2004emily}.
However, none of the prior works focus on the interplay between different sources of gender biases (e.g., ranking algorithms, \jobcontexts, and \candidate profiles) in online hiring. Our work makes the first attempt at understanding this interplay and analyzing its effect on hiring decisions. \\
\\
\textbf{Fair Ranking Algorithms}
Our work most closely resembles \citet{geyik2019fairness}, which seeks to understand the empirical effects of satisfying a \emph{ranked group fairness criterion}.
The ranked group fairness criterion as developed in \citet{zehlike2017fa} satisfies the properties that at any position in the ranking: 1) all groups are proportionally represented, 2) the relevance of the ranking is maximal subject to this constraint, and 3) within any group, candidates are of decreasing relevance. \citet{celis2020interventions} further study the theoretical guarantees of such ranking constraints. \citet{geyik2019fairness} conduct an A/B test on LinkedIn data using a post-hoc fairness re-ranking algorithm (\emph{Det-Greedy}) that ensures a desired proportional representation in top-ranked positions by greedily selecting the most relevant candidate available at each position in the ranking while maintaining maximum and minimum representation constraints for each group. In this way, \textit{Det-Greedy} generalizes the FA*IR algorithm developed in \cite{zehlike2017fa}, allowing for multiple protected groups and arbitrary distribution requirements. 
While \textit{Det-Greedy} was empirically evaluated,
the authors analyze the effectiveness of the re-ranking only with respect to specific business metrics but not equity of outcomes. 
On the contrary, our work exclusively focuses on analyzing gender-based disparities in online hiring decisions. 

Other works which focus on non-static rankings  optimize more detailed fairness criteria over a series of rankings. \citet{biega2018equity} optimize individual-level equity of attention, a measure of whether or not cumulative attention is proportional to cumulative relevance, amortized over successive rankings in which a candidate does not always appear at the same position.  \citet{singh2018fairness} optimize group fairness of exposure over a probabilistic distribution of rankings. 
Furthermore, learning to rank algorithms were proposed to ensure that fairness constraints are satisfied throughout the policy learning process, when relevance is not known a priori \cite{morik2020controlling, singh2019policy}. However, none of the aforementioned works carry out user studies to evaluate the effectiveness of the proposed algorithms in a real world setting. 
\\
\\
\textbf{User-Algorithm Interaction} 
Research on manipulated rankings finds that users have a strong bias toward the top items in a ranked list~\cite{keane2008people}. \citet{joachims2017accurately} attribute this effect partially to trust in the system generating the rankings, although they also find that item relevance mediates the effect of ranking.  
A study of Amazon Mechanical Turk workers finds that algorithm users have a strong preference for \emph{demographic parity} as a measure of fairness and are likely to prioritize accuracy over fairness in high stakes situations~\cite{srivastava2019mathematical}, potentially reducing the effectiveness of fairness-promoting recommendations.
Further, it has been demonstrated that algorithms intended to increase objectivity can result in disparate outcomes when biased users have agency to accept or reject the algorithmic recommendations \cite{green2019disparate}. While the aforementioned works provide insights into how users make decisions when presented with algorithmic recommendations, these works do not account for the dynamics of online hiring settings which is the key focus of this work.

\section{Problem Formulation}\label{sec:rq}
Our goal is to leverage real human decisions to determine if and how gender biases are perpetrated in online hiring. We aim to understand the interplay between various factors contributing to gender biases in these settings such as ranking algorithms, \candidate (e.g., worker, employee) profiles, nature of the jobs etc. We also want to examine the effectiveness of different kinds of ranking algorithms (including fair ranking algorithms) in these settings. More specifically, the goal of this work is to find answers to the following three critical questions: 
\begin{itemize}
    \item \textbf{RQ 1:} Do \employers exhibit gender bias uniformly across different \jobcontexts, \candidates, and rankings? Is gender bias universal or tied to traditional gender roles?  
    \item \textbf{RQ 2:} How effective are different ranking algorithms at mitigating gender bias? Are they equally effective in all settings, or do \candidate profiles and \jobcontexts impact their effectiveness? 
    \item \textbf{RQ 3:} Can fair ranking algorithms lead to disparate outcomes for different underrepresented groups (e.g., underrepresented groups comprising of males vs. females)?
\end{itemize}

In order to investigate the questions above, we obtain details of candidates (e.g., workers, employees), \jobcontexts, and candidate ranking data from TaskRabbit. In doing so, we are sampling directly from the data distribution of the TaskRabbit platform which in turn implies that the ranking data is possibly influenced by the feedback loops of the platform.
These feedback loops are an important source of bias that platforms seek to mitigate and are often difficult to simulate. Furthermore, utilizing data from TaskRabbit allows us to evaluate the efficacy of various ranking algorithms (including fair ranking algorithms) on real instances. While we conduct this study with data from TaskRabbit platform, we believe that our findings and insights are generic enough to be applicable to other online hiring platforms (e.g., Fiverr, Thumbtack, Upwork) which share similar characteristics. 
\section{Study Design}\label{sec:study-design}

In this section, we discuss in detail the design and execution of our study. First, we describe how we collect data from TaskRabbit. Next, we provide details about the candidates (workers, employees) in the data and describe the ranking algorithms we use in our study. Lastly, we conclude this section by describing how we simulate the online hiring setting in a user study with crowd workers (serving as proxy employers) from Amazon Mechanical Turk. 
\subsection{Data Collection}\label{subsec:data-collection}
We collect data from TaskRabbit by issuing the following three queries: \textbf{Shopping}, \textbf{Event Staffing}, and \textbf{Moving Assistance}. We chose these job categories because prior work~\cite{hannak2017bias} demonstrated that these categories capture varying levels of bias in favor of male workers. The authors found that while the Shopping category is highly biased in favor of male workers, the Moving category exhibits the least bias in their favor. \footnote{No studied contexts favored female workers.}

We collect only the top 10 results returned for each query. This is in line with the design of our UI (Figure 1) which displays only 10 candidates per query in a single page, thus eliminating the need for scrolling (more details later in this section). The geographic location corresponding to each query (e.g., moving assistance in NY) was varied so that different sets of individuals showed up in the query results. We identified the gender of the \candidates manually through their profile pictures and pronouns used in their descriptions and reviews. 

The geographic location associated with each query was refined until the returned ranking list comprised of 3 female candidates among the top 10 with most of them appearing in the bottom 5.
Excluding ranked lists with more than one female \candidate in the top 5 positions helped us ensure that there is scope for fair ranking algorithms to make the lists \emph{fairer}. If the top 5 positions in the lists are already heavily populated with females, the lists are already fair to begin with and fair ranking algorithms would have little to no impact. 
Furthermore, excluding ranked lists in which fewer than three women appear in the top 10 ensured that the application of fair ranking algorithms would not require us to substantially change the set of available candidates (i.e. add new \underrepresented \candidates). We note that our objective is not to retrieve a data set representative of TaskRabbit candidate (worker) rankings, but rather to sample data corresponding to real world scenarios in which fair ranking algorithms may benefit \underrepresented workers.
This process leaves us with three sets of 10 \candidates each where each set comprises of 7 male and 3 female \candidates. We denote these 3 sets as \textbf{D1}, \textbf{D2}, and \textbf{D3}. Each of these sets is a result of querying for one of the following job contexts (at different geographic locations) on TaskRabbit: shopping, event staffing, and moving assistance. Therefore, the sets differ from each other w.r.t. their feature distributions.
More details about the characteristics of each of these sets are provided in the following section.

\begin{table*}[t!]
\small
\centering
  \captionsetup{justification=centering}
\resizebox{\textwidth}{!}{
\begin{tabular}{|c|c|c|c|c|c|c|c|c|c|c|c|}
\hline
\cellcolor{ba}\textbf{} & \cellcolor{ba}\textbf{Rank} & \cellcolor{ba}\textbf{1} & \cellcolor{ba}\textbf{2} & \cellcolor{ba}\textbf{3} & \cellcolor{ba}\textbf{4} & \cellcolor{ba}\textbf{5} & \cellcolor{ba}\textbf{6} & \cellcolor{ba}\textbf{7} & \cellcolor{ba}\textbf{8} & \cellcolor{ba}\textbf{9} & \cellcolor{ba}\textbf{10} \\ \hline
\multirow{4}{*}{\textbf{D1}} & \cellcolor{ba}\textbf{Gender} & m & m & m & f & m & m & f & m & m & f \\ \cline{2-12} 
 & \cellcolor{ba}\textbf{\# Tasks Completed} & 634 & 395 & 64 & 41 & 158 & 388 & 141 & 458 & 48 & 7 \\ \cline{2-12} 
 &\cellcolor{ba} \textbf{\% Positive Reviews} & 99\% & 97\% & 100\% & 100\% & 100\% & 98\% & 100\% & 100\% & 100\% & 99\% \\ \cline{2-12} 
 & \cellcolor{ba}\textbf{\% Reliable} & 100\% & 100\% & 100\% & 100\% & 100\% & 100\% & 100\% & 98\% & 100\% & 100\% \\ \cline{2-12} 
 & \cellcolor{ba}\textbf{Relevance Score} & 0.8686 & 0.8590 & 0.8502 & 0.8485 & 0.8481 & 0.8467 & 0.8395 & 0.8384 & 0.8372 & 0.8339 \\ \specialrule{.1em}{.1em}{.1em}
\multirow{4}{*}{\textbf{D2}} & \cellcolor{ba}\textbf{Gender} & m & m & m & m & m & f & m & f & m & f \\ \cline{2-12} 
 & \cellcolor{ba}\textbf{\# Tasks Completed} & 0 & 1 & 0 & 0 & 6 & 10 & 2 & 2 & 2 & 3 \\ \cline{2-12} 
 & \cellcolor{ba}\textbf{\% Positive Reviews} & 95\% & 98\% & 100\% & 100\% & 100\% & 98\% & 97\% & 93\% & 99\% & 100\% \\ \cline{2-12} 
 & \cellcolor{ba}\textbf{\% Reliable} & 96\% & 100\% & 78\% & 100\% & 100\% & 100\% & 100\% & 100\% & 100\% & 100\% \\ \cline{2-12} 
 & \cellcolor{ba}\textbf{Relevance Score} & 0.7996 & 0.7882 & 0.7806 & 0.7711 & 0.7423 & 0.7412 &0.7376 & 0.7259 & 0.7195 & 0.7030 \\ \specialrule{.1em}{.1em}{.1em}
\multirow{4}{*}{\textbf{D3}} & \cellcolor{ba}\textbf{Gender} & m & m & m & m & m & m & m & f & f & f \\ \cline{2-12} 
 & \cellcolor{ba}\textbf{\# Tasks Completed} & 1 & 1 & 2 & 0 & 0 & 0 & 0 & 0 & 0 & 0 \\ \cline{2-12} 
 & \cellcolor{ba}\textbf{\% Positive Reviews} & 98\% & 96\% & 88\% & 100\% & 100\% & 100\% & 100\% & 100\% & 100\% & 100\% \\ \cline{2-12} 
 & \cellcolor{ba}\textbf{\% Reliable} & 100\% & 100\% & 100\% & 100\% & 100\% & 96\% & 100\% & 100\% & 83\% & 95\% \\ \cline{2-12} 
 & \cellcolor{ba}\textbf{Relevance Score} & 0.7771 & 0.7764 & 0.7241 & 0.7237 & 0.7190 & 0.6949 & 0.6912 & 0.6889 & 0.6795 & 0.6792 \\ \hline
\end{tabular}
}
\caption{
Sets D1, D2, and D3 collected from TaskRabbit with all the features used in this study. Ranks correspond to the rank assigned by the \rabbit algorithm.}\label{tbl:ranking-datasets}
\end{table*}
\subsection{Worker Features \& Data Description}
We extracted the  following features for each of the candidates (i.e., workers or employees): \textit{number of completed tasks}, \textit{\% positive reviews} and \textit{\% reliable}. 
Each candidate's original name was replaced with a different first-last name combination obtained from the most common white first names~\cite{firstnames} and last names~\cite{lastnames}. We do this in order to eliminate the effect of any racial or ethnic biases.

TaskRabbit shows additional features which are unrelated to the relevance of candidate (worker) features, such as price per hour, ``Elite Tasker" and ``Great Value" tags, recent reviews and profile descriptions. We chose to exclude these features as they may confound the relationship between feature relevance and selection probability.
Table \ref{tbl:ranking-datasets} captures the exact details of each of the sets D1, D2, and D3. We observe large differences in the values of the feature \textit{number of tasks completed} by \textit{gender} across the different sets. The disparities in the TaskRabbit relevance scores are smallest for D1 and largest for D3 with little within-group variation.
Below, we discuss the primary characteristics of each of the sets:
\begin{itemize}
    \item \textbf{D1}: In this set, the overrepresented \candidates \footnote{Throughout most of our analysis, the overrepresented \candidates will be men, as naturally occurs in the TaskRabbit queries. We later study the effect of fair ranking in situations in which traditional gender imbalances are reversed. In these cases, we will explicitly refer to the ranking as ``swapped", or {\sc (F$\leftrightarrow$M)}. In this case, women are the overrepresented group.} have completed substantially more tasks than the \underrepresented \candidates. The percentage of positive reviews and reliability are approximately equal across the two groups. 
    \item \textbf{D2}: In this set, \candidates have only completed a few tasks, with only one candidate each from the overrepresented group and the \underrepresented group having $>5$ tasks completed. All of the \underrepresented candidates score high on percentage of positive reviews and reliability, while there is more variance in the overrepresented group.
    \item \textbf{D3}: In this set, none of the \underrepresented \candidates have completed any tasks at all, and the overrepresented \candidates have completed between 0 and 2 tasks each. Several \candidates with no tasks completed 
    have maximum scores for the other features. The \candidate who has the most tasks completed (2) also has the lowest percentage of positive reviews. 
\end{itemize}

\begin{figure}[]
  \centering
  \captionsetup{justification=centering}
    \includegraphics[width=\columnwidth]{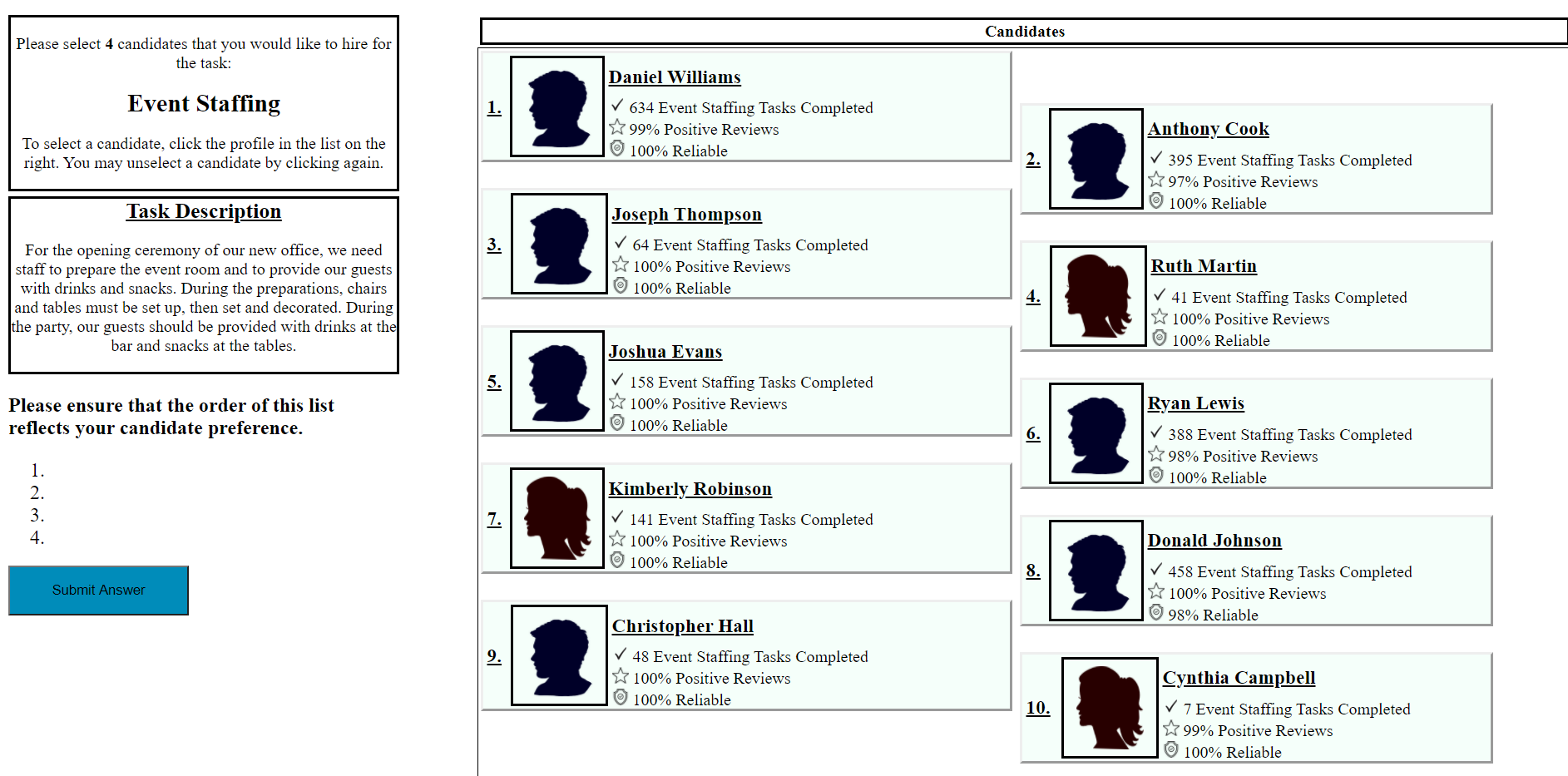}
        
        \caption{User interface using which participants (proxy employers) select job candidates (workers).}
    \label{fig:ranking-ui}
\end{figure}
\subsection{Ranking Algorithms} \label{sec:rankingalgos}
We leverage the following three ranking algorithms to rank order candidates (workers) as part of our study:
\begin{itemize}
    \item \rabbit : Candidates ranked by TaskRabbit relevance scores
    \item \random : Candidates ranked in random order
    \item \fair : Candidates ranked by Det-Greedy \cite{geyik2019fairness} applied as a post-processing step to TaskRabbit relevance scores
\end{itemize}
\fair requires two parameters to generate a fair ranking of any given set of \candidates: the proportion of male ($p_{male}$) and female ($p_{female}$) candidates in the underlying data distribution. We set $p_{male}=0.58$ and $p_{female}=0.42$ in all our experiments as this is in line with the actual gender distribution on the TaskRabbit platform~\cite{hannak2017bias}. 

To study whether there are disparate effects of fair ranking algorithms across genders, we also create a new version of each ranked list in which all the data remains the same but genders are swapped from female to male and vice versa. We denote the corresponding ranked lists with swapped genders as \rabbitswapped, \randomswapped and \fairswapped.
\subsection{Survey Design}\label{subsec:survey-design}
In order to analyze gender biases in online hiring decisions and determine how factors such as \jobcontexts, candidate (worker) features, and ranking algorithms influence these biases, we conduct a large scale user study with 1,079 participants on Amazon Mechanical Turk\footnote{All experiments were approved by our university's IRB.}. All participants were asked to perform three different ranking tasks, each one corresponding to a different job context (Shopping, Event Staffing, or Moving Assistance). During each ranking task, participants
were shown a ranked list of 10 \candidates and asked to select, in ranked order, their top four \candidate choices (see Figure \ref{fig:ranking-ui}). 
\paragraph{Briefing:}
Participants began the study by viewing a detailed description of the instructions of the task. Each participant was then asked to answer three comprehension check questions to ensure that they had clearly understood the objectives of the task. 
Participants were not allowed to proceed until they answered these questions correctly. Participants were told that if the company hired at least one candidate recommended by the participant, they would receive a bonus of $\$ 0.15$ (every participant received this bonus). We introduced the bonus in order to motivate the participants to think about their decisions, carefully consider the provided information, and not just click through as quickly as possible. 
Participants were also told that \candidates provide a full resume to the platform and a computer algorithm analyzes this information and ranks \candidates \textit{``according to various criteria including how likely they are to be hired and successfully complete the task."} 
This instruction was meant to suggest to the participants that the underlying ranking algorithm may be leveraging additional information (which is unavailable to the participants) when estimating relevance of the candidates. This, in turn, incentivizes the participants to take into account the ranking of the algorithm when making their own selections.
\paragraph{UI for Job Candidate Selection:}
Figure \ref{fig:ranking-ui} shows the responsive UI that we developed to display all 10 candidates in a single page eliminating the need for scrolling. 
Prior research has demonstrated that some users display an unwillingness to advance beyond initial results in UI designs involving infinite scrolling or pagination~\cite{o2006modeling}. This, in turn, can introduce other confounding factors (e.g., participants who never see candidates beyond page 1) which make it harder to accurately estimate the effect of gender biases in candidate selection. By removing scrolling effort, it is significantly more likely that all the candidates in the ranked list will be examined by the participants (proxy employers). Thus, we choose to display only 10 candidates in a single page. 
\paragraph{Job Candidate Selection:}
Participants then interacted with the simulated job \candidate selection process shown in Figure \ref{fig:ranking-ui}. In each of the three different \jobcontexts, \employers were asked to select 4 \candidates, in order of preference, to recommend to a company.
A task description was displayed to limit the possible interpretations of the tasks.
\begin{figure*}[h]
  \centering
  \captionsetup{justification=centering}
    \includegraphics[scale=0.35]{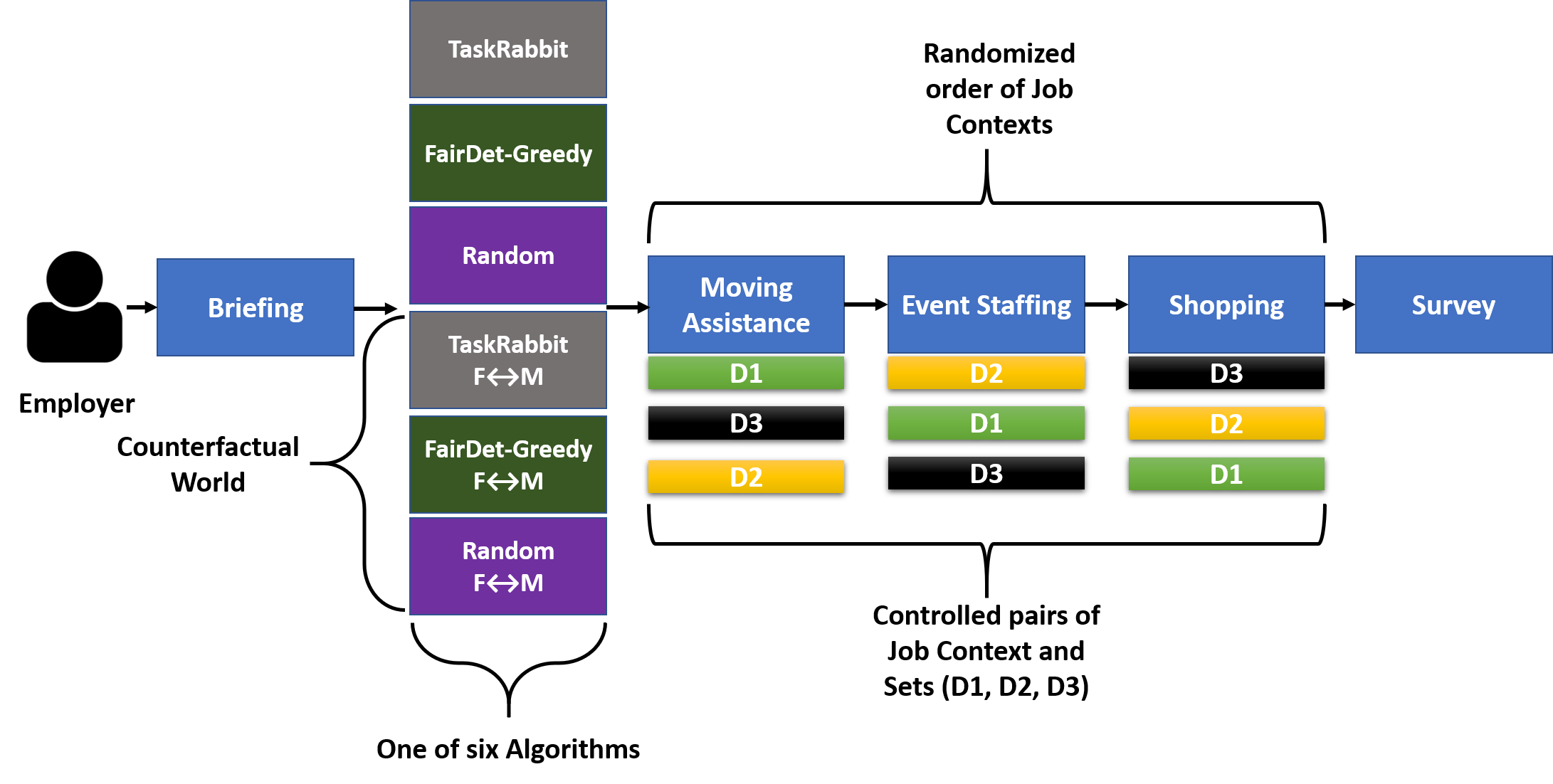}
    \caption{Visualization of our experimental setup. Participants are randomly assigned to one of the six ranking algorithms and see a random ordering of the \jobcontexts while we ensure each set (D1/D2/D3) appears exactly once alongside one of the \jobcontexts.}
    \label{fig:experiment_setup}
\end{figure*}
Participants were randomly assigned to one of the three ranking algorithms in either the setting in which \candidates have the same gender as in the original dataset obtained from TaskRabbit or the setting in which genders of the \candidates were swapped (See Section \ref{sec:rankingalgos}). 
Participants completed three ranking tasks: one for each \jobcontext. The order in which participants encountered different \jobcontexts was randomized. 
Each participant was shown all the three sets D1, D2, and D3 exactly once and each set was randomly paired with one of the three \jobcontexts (See Figure \ref{fig:experiment_setup}). 

For example, participant $A$ sees three ranked lists (one for each task) ordered by the \rabbit algorithm. First, they see a ranked list of candidates from set D2 alongside the \jobcontext of moving assistance. Next, they see another ranked list of candidates corresponding to set D3 with the job context event staffing, and lastly they see candidates from set D1 with job context shopping.
Participant $B$ might be randomly assigned to the algorithm \fair where they selects candidates ranked by \fair from D3 for shopping, from D1 for moving assistance, and then from D2 for event staffing. 
 
\paragraph{Additional Survey Questions}
After completing the job candidate selection process, we asked each participant to rate the importance of each of the displayed features for each job context on a 5-point Likert scale. We also asked the participants how much they trusted the computer system's assessment of the candidates on a 5-point Likert scale. We then asked the participants to describe their decision-making process with at least 40 characters in a free text field. Participants optionally self-reported gender, age, education level, and household income.
\subsection{Participant Demographics \& Compensation}
\begin{figure}[h]
  \centering
  \captionsetup{justification=centering}
    \includegraphics[scale=0.35]{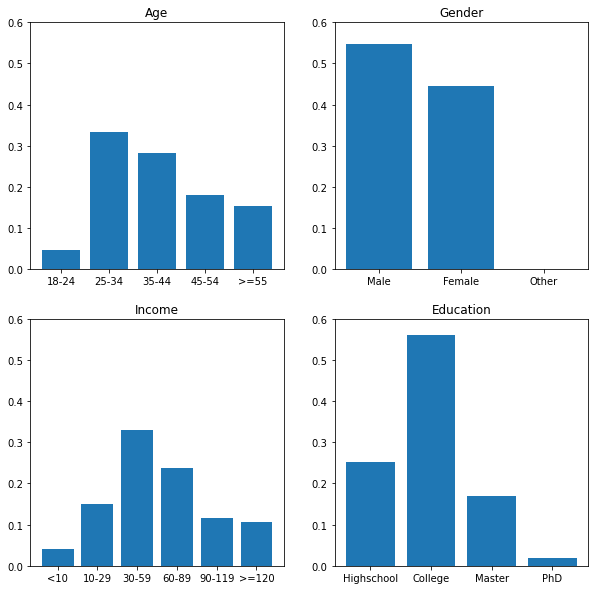}
    \caption{Distribution of age, gender, education and household income variables for participants (proxy \employers) in our study.}
    \label{fig:demographics}
\end{figure}

We recruited participants who can serve as proxy employers for our study on Amazon Mechanical Turk. All our participants were Turkers living in the US who had at least 5,000 approved tasks with $95\%$ approval rate. We compensated all the participants with a base rate of \$0.70 and a bonus of \$0.15. Every approved participant received the bonus. We disapproved someone only if they did not attempt to provide reasonable answers to the text field of the survey. With an average completion time of 6 minutes, we paid an average hourly wage of \$8.50. Out of the  1,079 participants in our study, $55\%$ identified themselves as male and $45\%$ as female. As seen in Figure \ref{fig:demographics}, $61\%$ of the participants were between $25$ and $44$ years old, $5\%$ were between $18$ and $24$ years old, and $15\%$ were older than $54$ years. The highest level of education was a high school degree for $25\%$ of the participants, a college degree for $56\%$, and a masters or PhD in case of the remaining $19\%$ of the participants. The median household income was between $\$30,000 - \$59,000$.
\section{Analysis \& Results}\label{sec:results}
In this section, we analyze the responses of all the 1,079 participants recruited on Amazon Mechanical Turk to find answers to the research questions highlighted in Section 3. First, we analyze whether participants (proxy employers) exhibit gender bias uniformly across different job contexts, candidates, and rankings. Then, we focus on assessing the effectiveness of different ranking algorithms including \fair in mitigating gender biases. Lastly, we examine whether fair ranking algorithms lead to disparate outcomes for different underrepresented groups. 
\hideh{
We find that:
\begin{itemize}
    \item Our \employers do exhibit pro-male gender bias after controlling for \candidate features and ranking, however this bias is only significant for our traditionally male-dominated \jobcontext (moving). 
    \item Controlling for visibility (all \candidates displayed in a single page), upranking \underrepresented gender \candidates increases the probability of selection. The size of this effect is mediated by \candidate and job features. 
    \item \Underrepresented men are selected at a higher rate than equivalent \underrepresented women. Most of this effect comes specifically from the traditionally male-dominated \jobcontext (moving). We do not find statistically significant evidence that fair ranking induces additional disparate effects for \underrepresented \candidates based on their gender.
\end{itemize}
In analyzing \employer bias and the effectiveness of fair ranking algorithms, we focus only on the data in which male \candidates tend to be ranked higher than female \candidates. Because the relevance scores are provided by TaskRabbit, we cannot know whether flipping the gender of the \candidates breaks correlations that are relevant to the ranking algorithm, e.g. through implicit feedback. For this reason, we reserve the data with flipped genders for examining disparate effects of \underrepresentation and fair ranking.}
\begin{table}[]
\centering
\captionsetup{justification=centering}
\resizebox{\columnwidth}{!}{%
\begin{tabular}{ccccc}
 &
  \multicolumn{2}{c}{\textbf{\begin{tabular}[c]{@{}c@{}}Candidate Selected\\ (w/o Interactions)\end{tabular}}} &
  \multicolumn{2}{c}{\textbf{\begin{tabular}[c]{@{}c@{}}Candidate Selected\\ (w/ Interactions)\end{tabular}}} \\
\multicolumn{1}{l}{} &
  \multicolumn{1}{l|}{4 Selections} &
  \multicolumn{1}{l|}{1 Selection} &
  \multicolumn{1}{l|}{4 Selections} &
  \multicolumn{1}{l}{1 Selection} \\ 
  \multicolumn{1}{l}{} &
  \multicolumn{1}{c|}{$(k=4)$} &
  \multicolumn{1}{c|}{$(k=1)$} &
  \multicolumn{1}{c|}{$(k=4)$} &
  \multicolumn{1}{c}{$(k=1)$} \\ \hline
(Intercept) &
  \multicolumn{1}{c|}{\textbf{-0.670***}} &
  \multicolumn{1}{c|}{\textbf{-2.316***}} &
  \multicolumn{1}{c|}{\textbf{-0.382***}} &
  \textbf{-1.170***} \\
Moving Assistance &
  \multicolumn{1}{c|}{-0.005} &
  \multicolumn{1}{c|}{0.025} &
  \multicolumn{1}{c|}{0.276} &
  0.561 \\
Shopping &
  \multicolumn{1}{c|}{-0.007} &
  \multicolumn{1}{c|}{0.014} &
  \multicolumn{1}{c|}{-0.085} &
  0.053 \\
Positive Reviews &
  \multicolumn{1}{c|}{\textbf{0.244***}} &
  \multicolumn{1}{c|}{\textbf{0.155***}} &
  \multicolumn{1}{c|}{\textbf{0.151***}} &
  0.170 \\
Reliability &
  \multicolumn{1}{c|}{\textbf{1.230***}} &
  \multicolumn{1}{c|}{\textbf{1.570***}} &
  \multicolumn{1}{c|}{\textbf{1.320***}} &
  \textbf{2.255***} \\
Completed Tasks &
  \multicolumn{1}{c|}{\textbf{0.215***}} &
  \multicolumn{1}{c|}{\textbf{0.281***}} &
  \multicolumn{1}{c|}{\textbf{0.229***}} &
  \textbf{0.291***} \\
Rank &
  \multicolumn{1}{c|}{\textbf{-0.226***}} &
  \multicolumn{1}{c|}{\textbf{-0.716***}} &
  \multicolumn{1}{c|}{\textbf{-0.066***}} &
  \textbf{-0.252***} \\
Female &
  \multicolumn{1}{c|}{\textbf{0.171***}} &
  \multicolumn{1}{c|}{-0.071} &
  \multicolumn{1}{c|}{0.003} &
  -0.691 \\ \hline
Female + Moving Assistance &
  \multicolumn{1}{c|}{} &
  \multicolumn{1}{c|}{} &
  \multicolumn{1}{c|}{\textbf{-0.385***}} &
  -0.300 \\
Female + Positive Reviews &
  \multicolumn{1}{c|}{} &
  \multicolumn{1}{c|}{} &
  \multicolumn{1}{c|}{\textbf{0.545***}} &
  \textbf{0.223***} \\
Female + Completed Tasks &
  \multicolumn{1}{c|}{} &
  \multicolumn{1}{c|}{} &
  \multicolumn{1}{c|}{\textbf{-0.751***}} &
  \textbf{-1.510***} \\
 &
   &
  \multicolumn{1}{l}{} &
  \multicolumn{1}{l}{} &
  
\end{tabular}%
}
\caption{Coefficients of a logistic regression 
model which predicts whether a \candidate will be selected as one of the top $k$ \candidates. Input features are \candidate features (See Table 1), \jobcontext, and gender. This model captures pairwise feature interactions. For readability, we only show interactions that are either of interest and/or are statistically significant (in bold). Note that the Intercept represents baseline regression coefficient for a Male \candidate being selected in the Event Staffing \jobcontext. \textit{$*** = p<\frac{0.001}{4} ; **= p<\frac{0.01}{4} ; *=\frac{0.05}{4}$}.
}
\label{tbl:logreg-employer-bias}
\end{table}
\begin{figure}[h]
     \centering
     \captionsetup{justification=centering}
     \begin{subfigure}{\columnwidth}
     \captionsetup{justification=centering}
         \centering
         \includegraphics[width=\columnwidth]{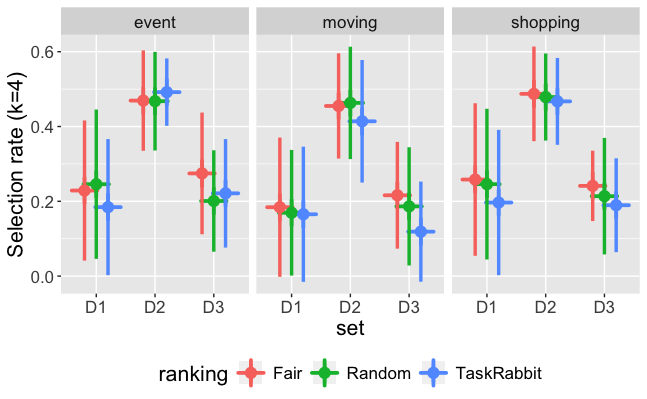}
    \caption{Selection rate of women, when women are the \underrepresented group.}
    \label{fig:meanmwselection}
    \end{subfigure}
    \begin{subfigure}{\columnwidth}
    \captionsetup{justification=centering}
         \centering
         \includegraphics[width=\columnwidth]{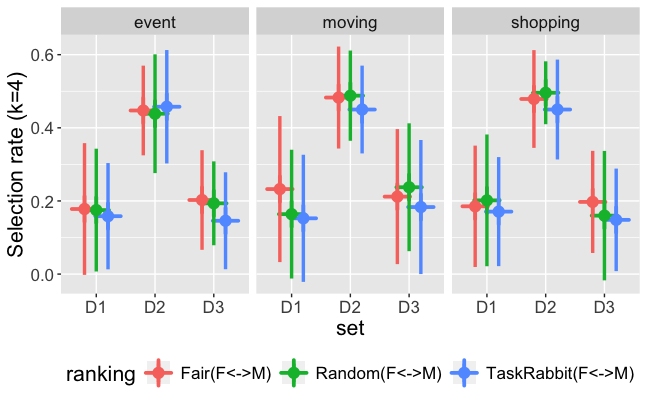}
    \caption{Selection rate of men, when men are the \underrepresented group.}
    \label{fig:meanwmselection}
    \end{subfigure}
    \caption{Selection rates of \underrepresented \candidates by different ranking algorithms across various \jobcontexts and sets D1,D2, D3. \fair algorithm consistently favors \underrepresented \candidates compared to \rabbit algorithm.
    }
    \label{fig:selections}
\end{figure}
\begin{table*}[t!]
\centering
\captionsetup{justification=centering}
\resizebox{\textwidth}{!}{%
\begin{tabular}{ccccccccc}
 &
  \textbf{\begin{tabular}[c]{@{}c@{}}Fraction of underrepresented \\ candidates per selection\\ (w/o Interactions)\end{tabular}} &
   &
   &
   &
  \textbf{\begin{tabular}[c]{@{}c@{}}Fraction of underrepresented \\ candidates per selection\\ (w/ Interactions)\end{tabular}} &
  \multicolumn{1}{l}{} &
  \multicolumn{1}{l}{} &
  \multicolumn{1}{l}{} \\ \hline
 &
  \multicolumn{1}{c|}{4 Selections} &
  \multicolumn{1}{c|}{3 Selections} &
  \multicolumn{1}{c|}{2 Selections} &
  \multicolumn{1}{c|}{1 Selection} &
  \multicolumn{1}{c|}{4 Selections} &
  \multicolumn{1}{l|}{3 Selections} &
  \multicolumn{1}{l|}{2 Selections} &
  \multicolumn{1}{l}{1 Selection} \\ 
  &
    \multicolumn{1}{c|}{$(k=4)$} &
  \multicolumn{1}{c|}{$(k=3)$} &
  \multicolumn{1}{c|}{$(k=2)$} &
  \multicolumn{1}{c|}{$(k=1)$} &
  \multicolumn{1}{c|}{$(k=4)$} &
  \multicolumn{1}{c|}{$(k=3)$} &
  \multicolumn{1}{c|}{$(k=2)$} &
  \multicolumn{1}{c}{$(k=1)$} \\ \hline
(Intercept) &
  \multicolumn{1}{c|}{\textbf{-0.780***}} &
  \multicolumn{1}{c|}{\textbf{-0.632***}} &
  \multicolumn{1}{c|}{\textbf{-0.891***}} &
  \multicolumn{1}{c|}{\textbf{-1.020***}} &
  \multicolumn{1}{c|}{0.029} &
  \multicolumn{1}{c|}{\textbf{0.463***}} &
  \multicolumn{1}{c|}{0.270} &
  0.373 \\

\rabbit &
  \multicolumn{1}{c|}{\textbf{-0.183***}} &
  \multicolumn{1}{c|}{\textbf{-0.283***}} &
  \multicolumn{1}{c|}{\textbf{-0.373***}} &
  \multicolumn{1}{c|}{\textbf{-0.633***}} &
  \multicolumn{1}{c|}{-0.097} &
  \multicolumn{1}{c|}{-0.195} &
  \multicolumn{1}{c|}{\textbf{0.443*}} &
  \textbf{-0.964**} \\
FLIP &
  \multicolumn{1}{c|}{\textbf{-0.091*}} &
  \multicolumn{1}{c|}{-0.080} &
  \multicolumn{1}{c|}{-0.129} &
  \multicolumn{1}{c|}{-0.008} &
  \multicolumn{1}{c|}{-0.228} &
  \multicolumn{1}{c|}{-0.203} &
  \multicolumn{1}{c|}{-0.410} &
  -0.396 \\
D2 &
  \multicolumn{1}{c|}{\textbf{-0.007}} &
  \multicolumn{1}{c|}{-0.081} &
  \multicolumn{1}{c|}{0.042} &
  \multicolumn{1}{c|}{0.032} &
  \multicolumn{1}{c|}{\textbf{-1.335***}} &
  \multicolumn{1}{c|}{\textbf{-1.801***}} &
  \multicolumn{1}{c|}{\textbf{-1.636***}} &
  \textbf{-2.209***} \\
D3 &
  \multicolumn{1}{c|}{0.032} &
  \multicolumn{1}{c|}{-0.081} &
  \multicolumn{1}{c|}{-0.025} &
  \multicolumn{1}{c|}{-0.104} &
  \multicolumn{1}{c|}{\textbf{-1.105***}} &
  \multicolumn{1}{c|}{\textbf{-1.827***}} &
  \multicolumn{1}{c|}{\textbf{-1.876***}} &
  \textbf{-2.246***} \\
Shopping &
  \multicolumn{1}{c|}{0.010} &
  \multicolumn{1}{c|}{-0.007} &
  \multicolumn{1}{c|}{0.130} &
  \multicolumn{1}{c|}{0.144} &
  \multicolumn{1}{c|}{\textbf{-1.281***}} &
  \multicolumn{1}{c|}{\textbf{-1.825***}} &
  \multicolumn{1}{c|}{\textbf{-1.760***}} &
  \textbf{-2.140***} \\
Moving &
  \multicolumn{1}{c|}{-0.075} &
  \multicolumn{1}{c|}{-0.075} &
  \multicolumn{1}{c|}{-0.035} &
  \multicolumn{1}{c|}{0.000} &
  \multicolumn{1}{c|}{\textbf{-1.496***}} &
  \multicolumn{1}{c|}{\textbf{-1.977***}} &
  \multicolumn{1}{c|}{\textbf{-2.118***}} &
  \textbf{-2.485***} \\ \hline
\begin{tabular}[c]{@{}c@{}}Moving + D2\end{tabular} &
  \multicolumn{1}{c|}{} &
  \multicolumn{1}{c|}{} &
  \multicolumn{1}{c|}{} &
  \multicolumn{1}{c|}{} &
  \multicolumn{1}{c|}{\textbf{1.326***}} &
  \multicolumn{1}{c|}{\textbf{1.497***}} &
  \multicolumn{1}{c|}{\textbf{1.389***}} &
  \textbf{2.275***} \\
\begin{tabular}[c]{@{}c@{}}Moving + D3\end{tabular} &
  \multicolumn{1}{l|}{} &
  \multicolumn{1}{l|}{} &
  \multicolumn{1}{l|}{} &
  \multicolumn{1}{l|}{} &
  \multicolumn{1}{c|}{\textbf{2.495***}} &
  \multicolumn{1}{c|}{\textbf{3.559***}} &
  \multicolumn{1}{c|}{\textbf{3.841***}} &
  \textbf{4.729***} \\
\begin{tabular}[c]{@{}c@{}}Moving + FLIP\end{tabular} &
  \multicolumn{1}{l|}{} &
  \multicolumn{1}{l|}{} &
  \multicolumn{1}{l|}{} &
  \multicolumn{1}{l|}{} &
  \multicolumn{1}{c|}{\textbf{0.396**}} &
  \multicolumn{1}{c|}{\textbf{0.478*}} &
  \multicolumn{1}{c|}{\textbf{0.733*}} &
  0.504 \\
\begin{tabular}[c]{@{}c@{}}Shopping + D2\end{tabular} &
  \multicolumn{1}{l|}{} &
  \multicolumn{1}{l|}{} &
  \multicolumn{1}{l|}{} &
  \multicolumn{1}{l|}{} &
  \multicolumn{1}{c|}{\textbf{2.652***}} &
  \multicolumn{1}{c|}{\textbf{3.607***}} &
  \multicolumn{1}{c|}{\textbf{3.458***}} &
  \textbf{4.138***} \\
\begin{tabular}[c]{@{}c@{}}Shopping + D3\end{tabular} &
  \multicolumn{1}{l|}{} &
  \multicolumn{1}{l|}{} &
  \multicolumn{1}{l|}{} &
  \multicolumn{1}{l|}{} &
  \multicolumn{1}{c|}{\textbf{1.247***}} &
  \multicolumn{1}{c|}{\textbf{2.191***}} &
  \multicolumn{1}{c|}{\textbf{2.261***}} &
  \textbf{2.233***} \\
\end{tabular}%
}
\caption{Coefficients of a linear regression model
which predicts the fraction of \underrepresented \candidates selected in the top $k$ set of candidates. The input features are ranking type, set, and \jobcontext. This model captures pairwise feature interactions. For readability, we only show interactions that are either of interest and/or are statistically significant (in bold). Note that the Intercept coefficient represents baseline fraction of underrepresented \candidates selected using \fair in the Event Staffing \jobcontext. \textit{$*** = p<\frac{0.001}{4} ; **= p<\frac{0.01}{4} ; *=\frac{0.05}{4}$}. \textbf{FLIP} is a variable that indicates if a given candidate is from the counterfactual setting where men are underrepresented (FLIP = 1) or not (FLIP = 0). 
}\label{tbl:linear-regression-rq3}
\end{table*}
\subsection{Employer Bias}\label{subsec:employer-bias}
To study whether participants (proxy employers) exhibit gender bias when selecting candidates after controlling for \candidate features and rank, we carry out the following analysis. We train a logistic regression model to predict whether a given candidate will be chosen as one of the top $k$ (for $k=\{1,2,3,4\})$ \candidates. This analysis considers only the data corresponding to actual gender distribution on the TaskRabbit platform, in which women are \underrepresented (with counterfactual data reserved for Section \ref{subsec:disparate-impact}.)  Features provided as input to this model include \candidate features (e.g., \% positive reviews, \% tasks completed, \% reliable), \candidate's rank and gender. Note that the above analysis will result in four different logistic regression models one for each value of $k$. The goal here is to examine the statistical significance of the feature coefficients (particularly that of the gender variable) of the resulting logistic regression models, which in turn allows us to determine which features are influential to the prediction.  We repeat the aforementioned analysis and build other logistic regression models which also include additional terms to capture the pairwise feature interactions between all the features listed above~\cite{logreg}. 

We use participant (proxy employer) rankings to study how gender impacts selections at 1, 2, 3, and 4 rank positions, assuming that the proxy \employer ranks first the \candidate they would have selected if they were only allowed to select one, and so on. Since we repeat this test 4 times, we apply a Bonferroni correction\footnote{https://mathworld.wolfram.com/BonferroniCorrection.html} to resulting p-values. Data is standardized such that all variables have a mean of 0 and standard deviation of 1, allowing for comparison of the coefficients. We additionally cluster the standard errors on participant IDs (i.e., mTurk WorkerIDs) to account for dependencies in our data and avoid over-reporting significance. 
Table~\ref{tbl:logreg-employer-bias} captures the results of this analysis for the cases where $k=1$ and $k=4$.

When $k=4$ and when we do not include pairwise feature interaction terms, the variable \emph{female} has a positive coefficient ($p < .001$), suggesting that female candidates are selected more often than might be expected. This may be due to proxy \employers consciously trying to enforce demographic parity across all of their selections (See Section~\ref{sec:employers_fairness} for more details). 
When pairwise feature interaction terms are included, we find that the variable \emph{female} is no longer significant across all tasks, but there is a significant negative coefficient on the interaction term comprising of the features \emph{female} and \emph{moving \jobcontext} ($p<.001$). This implies that females are less likely to be chosen for this job relative to event staffing jobs after controlling for features. 
We also find that females tend to benefit more from positive reviews compared to male \candidates ($p <.001$).

\subsection{Effectiveness of Fair Ranking} To study whether applying a post-hoc fair ranking algorithm helps mitigate gender biases in hiring decisions, we conduct a 3-way ANOVA with ranking type (\rabbit, \random, \fair \footnote{Again, swapped gender data is reserved for Section \ref{subsec:disparate-impact}.}), candidate set type (D1, D2, D3), and \jobcontext (moving assistance, event staffing, shopping) as the three factors, and \% of \candidates selected who are female as the dependent variable. 
We repeat this analysis for different values of $k$ i.e., $k= \{1, 2, 3, 4\}$. \footnote{This allows us to confirm whether or not the between-group variance is significantly greater than within-group variance before applying individual coefficient tests, reducing the probability of false discovery.} Since we repeat this test 4 times, we apply a Bonferroni correction to resulting p-values. 

We then compute the F-statistic both with and without considering the feature interaction terms. The F-statistic confirms that ranking type and candidate set type are statistically significant ($p<.001$) for all values of $k$, and \jobcontext is significant for all values of $k$ ($p<.001$) except for $k=1$. 

We conduct two more significance tests -- Wald test with clustered standard errors (on mTurk WorkerID) and Tukey's HSD  test\footnote{Tukey's Honest Significant Difference (HSD) test is a conservative post-hoc test which adjusts for the testing of multiple hypotheses. After an ANOVA finds significant differences, Tukey's HSD test can show which specific group means (compared pairwise) are the reason for the overall difference.}. These tests demonstrate that the difference between \rabbit and \fair is statistically significant across all values of $k$ when pairwise feature interactions are not modeled ($p<.001$). We find that \underrepresented \candidates obtain better outcomes under \fair. When we do include interaction terms, we find that the difference between \rabbit vs. \fair continues to be significant ($p<.001$) at all values of $k$ (all selection levels) according to Tukey's HSD and is only significant at $k=1$ and $k=2$ ($p<.05$ before correction) according to the Wald test. 

While we find that interaction terms are insignificant according to the Wald Test, exploratory analysis (see Figure \ref{fig:meanmwselection}) reveals that candidate set type (D1, D2, and D3) and \jobcontext both affect the efficacy of \fair. In particular, \fair appears to be most successful at increasing the proportion of \underrepresented \candidates selected in D3. Recall that in D3, the \underrepresented \candidates receive lower rankings but share similar feature values to overrepresented \candidates. In D1, \underrepresented \candidates have significantly fewer jobs completed relative to the overrepresented candidates, which appears to dampen the effectiveness of \fair relative to \rabbit. In D2, \underrepresented \candidates have more desirable feature values than their overrepresented counterparts and thus are already selected approximately at similar rates as their overrepresented counterparts. This appears to mitigate the effect of \fair, as \employers already try to consciously enforce demographic parity when they make decisions.

\begin{table*}[t]
\tiny
\centering
\captionsetup{justification=centering}
\resizebox{\textwidth}{!}{%
\begin{tabular}{|c|c|c|c|c|c|c|c|c|c|c|c|c|}
\hline
\multicolumn{13}{|c|}{\cellcolor{ba}\textbf{Moving assistance}}         \cellcolor{ba}                                                                                              \\ \hline
                   & \multicolumn{4}{c|}{\cellcolor{ba}\textbf{Task Rabbit}} & \multicolumn{4}{c|}{\cellcolor{ba}\textbf{Random}}  & \multicolumn{4}{c|}{\cellcolor{ba}\textbf{Fair}}    \\ \hline
\cellcolor{ba}\textbf{\#Choices} &\cellcolor{ba} \textbf{$k=1$}        &\cellcolor{ba} \textbf{$k=2$}        &\cellcolor{ba} \textbf{$k=3$}        &\cellcolor{ba} \textbf{$k=4$}        &\cellcolor{ba} \textbf{$k=1$}       &\cellcolor{ba} \textbf{$k=2$}       &\cellcolor{ba} \textbf{$k=3$}       &\cellcolor{ba} \textbf{$k=4$}       &\cellcolor{ba} \textbf{$k=1$}      &\cellcolor{ba} \textbf{$k=2$}       &\cellcolor{ba} \textbf{$k=3$}       &\cellcolor{ba} \textbf{$k=4$}      \\ \hline
\cellcolor{ba}\textbf{D1}        & 4.8\%   & 9.7\%   & 15.6\%  & 16.5\%  &\cellcolor{5-10} 11.3\% & \cellcolor{5-10}16.1\% &\cellcolor{<5} 16.1\% &\cellcolor{<5} 16.9\% &\cellcolor{5-10} 12.3\% & \cellcolor{5-10}16.7\% &\cellcolor{<5} 18.7\% &\cellcolor{<5}  18.4\% \\ \hline
\cellcolor{ba}\textbf{D2}        & 24.6\%  & 39.3\%  & 44.3\%  & 41.4\%  &\cellcolor{>15}  44.3\% &\cellcolor{5-10} 48.4\% &\cellcolor{5-10} 53.0\% & \cellcolor{<5}46.3\% &\cellcolor{>15}60.7\% &\cellcolor{5-10} 46.7\% &\cellcolor{5-10} 51.9\% &\cellcolor{<5} 45.5\% \\ \hline
\cellcolor{ba}\textbf{D3}        & 1.6\%   & 5.7\%   & 9.8\%   & 11.9\%  &\cellcolor{5-10} 10.2\% &\cellcolor{5-10} 11.9\% &\cellcolor{<5} 14.1\% &\cellcolor{5-10} 18.7\% &\cellcolor{5-10} 8.5\%  &\cellcolor{5-10} 14.4\% &\cellcolor{5-10} 17.0\% &\cellcolor{5-10} 21.6\% \\ \specialrule{.1em}{.1em}{.1em}
\cellcolor{ba}\textbf{all}       & 10.3\%  & 18.3\%  & 23.2\%  & 23.3\%  &\cellcolor{10-15} 22.0\% &\cellcolor{5-10} 25.5\% & \cellcolor{<5} 27.8\% & \cellcolor{<5} 27.3\% &\cellcolor{>15} 27.7\% & \cellcolor{5-10} 25.9\% &\cellcolor{5-10} 29.2\% &\cellcolor{5-10} 28.5\% \\ \hline
\cellcolor{ba}\textbf{FLIP-all}       & 17.3\%     & 22.6\%    & 27.1\%    & 26.2\%    & \cellcolor{<5}26.1\%   &\cellcolor{5-10} 28.2\%   &\cellcolor{<5} 30.8\%   &\cellcolor{<5} 29.7\%   &\cellcolor{5-10} 25.8\%   &\cellcolor{5-10} 28.5\%   &\cellcolor{5-10} 32.8\%  &\cellcolor{<5} 30.9\%  \\ \hline
\multicolumn{1}{l}{}                     & \multicolumn{1}{l}{}                    & \multicolumn{1}{l}{}                            & \multicolumn{1}{l}{}                            & \multicolumn{1}{l}{}                                   & \multicolumn{1}{l}{}                              & \multicolumn{1}{l}{}                                     \\ \hline
\multicolumn{13}{|c|}{\cellcolor{ba}\textbf{Event staffing}}                                                                                                 \\ \hline
\textbf{\cellcolor{ba}}          & \multicolumn{4}{c|}{\cellcolor{ba}\textbf{Task Rabbit}} & \multicolumn{4}{c|}{\cellcolor{ba}\textbf{Random}}  & \multicolumn{4}{c|}{\cellcolor{ba}\textbf{Fair}}    \\ \hline
\cellcolor{ba}\textbf{\#Choices} &\cellcolor{ba} \textbf{$k=1$}        &\cellcolor{ba} \textbf{$k=2$}        &\cellcolor{ba} \textbf{$k=3$}        &\cellcolor{ba} \textbf{$k=4$}        &\cellcolor{ba} \textbf{$k=1$}       &\cellcolor{ba} \textbf{$k=2$}       &\cellcolor{ba} \textbf{$k=3$}       &\cellcolor{ba} \textbf{$k=4$}       &\cellcolor{ba} \textbf{$k=1$}      &\cellcolor{ba} \textbf{$k=2$}       &\cellcolor{ba} \textbf{$k=3$}       &\cellcolor{ba} \textbf{$k=4$}      \\ \hline
\cellcolor{ba}\textbf{D1}        & 4.9\%   & 11.5\%  & 16.9\%  & 18.4\%  &\cellcolor{5-10} 10.2\% &\cellcolor{10-15} 22.9\% &\cellcolor{5-10} 24.3\% &\cellcolor{5-10} 24.6\% &\cellcolor{10-15} 17.0\% &\cellcolor{10-15} 22.0\% &\cellcolor{5-10} 22.6\% &\cellcolor{<5} 22.9\% \\ \hline
\cellcolor{ba}\textbf{D2}        & 37.1\%  & 46.8\%  & 58.1\%  & 49.2\%  &\cellcolor{10-15} 48.4\% &\cellcolor{<5} 48.4\% &\cellcolor{-5} 54.8\% &\cellcolor{-5} 46.8\% &\cellcolor{>15} 52.6\% &\cellcolor{<5} 50.9\% &\cellcolor{-5} 56.1\% &\cellcolor{-5} 46.9\% \\ \hline
\cellcolor{ba}\textbf{D3}        & 4.9\%   & 12.3\%  & 14.8\%  & 22.1\%  &\cellcolor{<5} 8.2\%  &\cellcolor{-5} 10.7\% &\cellcolor{-5} 13.1\% &\cellcolor{-5} 20.1\% & \cellcolor{10-15}16.4\% &\cellcolor{5-10} 20.5\% &\cellcolor{5-10} 23.5\% &\cellcolor{5-10} 27.5\% \\ \specialrule{.1em}{.1em}{.1em}
\cellcolor{ba}\textbf{all}       & 15.6\%  & 23.5\%  & 29.9\%  & 29.9\%  &\cellcolor{5-10} 22.3\% &\cellcolor{<5} 27.3\% &\cellcolor{<5} 30.8\% & \cellcolor{<5}30.5\% &\cellcolor{10-15} 28.7\% &\cellcolor{5-10} 31.1\% &\cellcolor{<5} 34.1\% &\cellcolor{<5} 32.4\% \\ \hline
\cellcolor{ba}\textbf{FLIP-all}       & 15.2\%     & 20.3\%    & 25.5\%    & 25.4\%    & \cellcolor{5-10}20.8\%   &\cellcolor{<5} 24.3\%   &\cellcolor{<5} 28.2\%   &\cellcolor{<5} 26.9\%   &\cellcolor{5-10} 22.1\%   &\cellcolor{<5} 23.4\%   &\cellcolor{<5} 29.7\%  &\cellcolor{<5} 27.6\%  \\ \hline
\multicolumn{1}{l}{}                     & \multicolumn{1}{l}{}                    & \multicolumn{1}{l}{}                            & \multicolumn{1}{l}{}                            & \multicolumn{1}{l}{}                                   & \multicolumn{1}{l}{}                              & \multicolumn{1}{l}{}                                     \\ \hline
\multicolumn{13}{|c|}{\cellcolor{ba}\textbf{Shopping}}                                                                                                       \\ \hline
\cellcolor{ba}\textbf{}          & \multicolumn{4}{c|}{\cellcolor{ba}\textbf{Task Rabbit}} & \multicolumn{4}{c|}{\cellcolor{ba}\textbf{Random}}  & \multicolumn{4}{c|}{\cellcolor{ba}\textbf{Fair}}    \\ \hline
\cellcolor{ba}\textbf{\#Choices} &\cellcolor{ba} \textbf{$k=1$}        &\cellcolor{ba} \textbf{$k=2$}        &\cellcolor{ba} \textbf{$k=3$}        &\cellcolor{ba} \textbf{$k=4$}        &\cellcolor{ba} \textbf{$k=1$}       &\cellcolor{ba} \textbf{$k=2$}       &\cellcolor{ba} \textbf{$k=3$}       &\cellcolor{ba} \textbf{$k=4$}       &\cellcolor{ba} \textbf{$k=1$}      &\cellcolor{ba} \textbf{$k=2$}       &\cellcolor{ba} \textbf{$k=3$}       &\cellcolor{ba} \textbf{$k=4$}       \\ \hline
\cellcolor{ba}\textbf{D1}        &
6.6\%   & 14.8\%  & 15.3\%  & 19.7\%  &\cellcolor{5-10} 16.4\% &\cellcolor{5-10} 23.0\% &\cellcolor{5-10} 23.5\% &\cellcolor{5-10} 24.6\% &\cellcolor{<5} 11.5\% &\cellcolor{10-15} 26.2\% & \cellcolor{10-15} 28.4\% &\cellcolor{5-10} 25.8\% \\ \hline
\cellcolor{ba}\textbf{D2}        & 47.5\%  & 49.2\%  & 52.5\%  & 46.7\%  & \cellcolor{<5} 49.2\% &\cellcolor{<5} 53.4\% &\cellcolor{<5} 57.1\% &\cellcolor{<5} 47.9\% &\cellcolor{5-10} 54.2\% &\cellcolor{5-10} 55.1\% &\cellcolor{<5} 57.1\% &\cellcolor{<5} 48.7\% \\ \hline
\cellcolor{ba}\textbf{D3}        & 8.1\%   & 11.3\%  & 13.4\%  & 19.0\%  & 8.1\%  &\cellcolor{<5} 12.1\%  &\cellcolor{<5} 17.2\%  &\cellcolor{<5} 21.4\% &\cellcolor{5-10} 17.5\% &\cellcolor{5-10} 20.2\% &\cellcolor{5-10} 22.8\% &\cellcolor{5-10} 24.1\% \\ \specialrule{.1em}{.1em}{.1em}
\cellcolor{ba}\textbf{all}       & 20.7\%  & 25.1\%  & 27.1\%  & 28.5\%  & \cellcolor{<5}24.5\% &\cellcolor{<5} 29.5\% &\cellcolor{5-10} 32.6\% &\cellcolor{<5} 31.3\% &\cellcolor{5-10} 27.8\% &\cellcolor{5-10} 33.8\% &\cellcolor{5-10} 36.1\% &\cellcolor{<5} 32.9\% \\ \hline
\cellcolor{ba} \textbf{FLIP-all}       & 16.1\%     & 20.0\%    & 24.5\%    & 25.6\%    & \cellcolor{5-10}24.3\%   &\cellcolor{5-10} 26.5\%   &\cellcolor{5-10} 30.3\%   &\cellcolor{<5} 28.6\%   &\cellcolor{5-10} 25.7\%   &\cellcolor{5-10} 28.5\%   &\cellcolor{5-10} 30.3\%  &\cellcolor{<5} 28.7\%  \\ \hline
\end{tabular}%
}
\caption{Percentage of female candidates selected in the top $k$ set of candidates by various ranking algorithms across various \jobcontexts and sets (D1, D2, and D3). 
Colors indicate the change in percentage points compared to
\rabbit: \colorbox{-5-10}{[-10,-5]} \colorbox{-5}{(-5,0)}|0|\colorbox{<5}{(0,+5)}|\colorbox{5-10}{[+5,+10)}|\colorbox{10-15}{[+10,+15)}|\colorbox{>15}{[+15,$\infty$)}\\
\textbf{all} captures the selection rates of female candidates across all three sets D1, D2, and D3.
\textbf{FLIP-all} shows the selection rates of male candidates across all three sets D1, D2, and D3 with swapped genders (male candidates as the underrepresented group). In this row, colors indicate the change in percentage points compared to \rabbitswapped.
\\
}
\label{tbl:algo-compare-selection-rate m>f}
\end{table*}
\subsection{Fine-grained Analysis of the Impact of Fair Ranking}\label{User bias versus algorithm bias}
In this section, we dive deeper and attempt to understand the impact of different ranking algorithms on gender biases in hiring decisions involving various \jobcontexts, candidate set types (D1, D2, D3),  and different values of $k$. Table \ref{tbl:algo-compare-selection-rate m>f} captures the results for this analysis. Each cell in Table \ref{tbl:algo-compare-selection-rate m>f} corresponds to a particular combination (e.g., \jobcontext = moving assistance, ranking algorithm = \rabbit, and set = D1, and $k=1$) of the aforementioned aspects and the value in the cell represents the percentage of female candidates selected by participants for that particular combination. 

Table \ref{tbl:algo-compare-selection-rate m>f} captures several interesting insights. For example, in case of set D1, \jobcontext of moving, and \rabbit algorithm, $16.53\%$ of all selected candidates were female while $4.84\%$ of all first selections were female. The row \textit{``all"} captures the selection rates of female candidates across all three sets D1, D2, and D3.
Our results reveal that \fair increases the representation of female candidates in almost all cells compared to \rabbit. \fair is particularly effective in increasing female representation at participants' (proxy employers') first selection in all job contexts . We find the highest increase of $17.35$ percentage points in the first selection in case of moving assistance followed by event staffing (an increase of $13.02$ percentage points). 
Across all \jobcontexts and sets (D1, D2, D3), we observe that the difference between \fair and \rabbit decreases as the number of selections increases, suggesting that \fair mainly pushes female candidates higher in the priority list of participants but has lesser impact on the overall fraction of the selected female candidates.

\subsection{Disparate Impact of Fair Rankings} \label{subsec:disparate-impact}
In this section, we study whether fair ranking algorithms lead to disparate outcomes for different underrepresented groups. To carry out this analysis, we additionally use the data from settings in which gender labels are swapped (\fairswapped, \rabbitswapped, \randomswapped, see Section \ref{sec:rankingalgos}), representing the \emph{counterfactual} world in which men are \underrepresented.
We then carry out a linear regression to predict the percentage of underrepresented candidates (female in the original data, male in the counterfactual data) selected. This model is trained on the original data from sets D1, D2, and D3, as well as the aforementioned counterfactual data. The input variables to this regression task are set type (D1, D2, D3), \jobcontext, type of ranking algorithm (\fair vs. \rabbit), and a variable called \textbf{FLIP} that indicates if the corresponding candidate is from the counterfactual data (FLIP = 1) or not (FLIP = 0). If identical \underrepresented female candidates and \underrepresented male candidates are treated equally, we expect to see no significant coefficients on the FLIP variable or any of its interactions. 

Table~\ref{tbl:linear-regression-rq3} captures the feature coefficients output by the aforementioned linear regression task. We find that the feature coefficient corresponding to the FLIP variable is negative and significant at $k=4$ (4 selections), suggesting that \underrepresented men are \emph{less} likely to be selected than their female counterparts. 
When pairwise feature interaction terms are modeled in the linear regression, we find that the coefficient corresponding to the interaction term involving moving assistance and FLIP is positive and significant. 
We find no significant interactions between FLIP and the choice of the ranking algorithm. 
Exploratory analysis (See Figure \ref{fig:meanwmselection}) reveals that \fair algorithm favors the selection of men (in the counterfactual case where men are underrepresented) for moving assistance jobs even when they appear to be underqualified relative to women, in D1. 
\section{Exploratory Analysis} In this section, we discuss the exploratory analysis that we carried out to find answers to the following key questions: 1) Which \employer demographic groups perpetrate hiring biases? 2) Do \employers actively apply fairness criteria when making hiring decisions?
These questions were inspired by the textual answers we received from the study participants when we asked them to describe their decision-making process (one of our survey questions). We believe that the insights obtained from these answers can help pave the way for future work at the intersection of algorithmic fairness and human-computer interaction.
\subsection{Gender Biases of Different Employer Groups} \label{sec:employers_bias}
Here, we study if different employer groups exhibit different kinds of biases in the context of online hiring. We consider all \jobcontexts and both the counterfactual as well as the real data. We carry out linear regression with clustered standard errors on participant IDs to predict the percentage of selected female \candidates at different values of $k$. The input variables to this regression task are gender, age , income and education of the participants who serve as proxy \employers in our study. However, our sample size is too small to capture effects of all possible feature combinations from our survey. Thus we divided the user features age, income and education in two groups each. We divided \textit{age} in younger/older than 35 years and \textit{income} in less/more than \$60k annual household income. Furthermore we divided the \textit{education} variable in participants holding not more than a high school degree and participants holding at least a bachelors degree. The variable \textit{isAcademic} is true for the latter one.
Table \ref{tbl:bias-demographics} shows the significant interactions of the aforementioned \employer features. Note that the intercept represents male \employers who are older than 35 years, have not more than a high school degree and an annual household income of more than \$60k. 
We observe that only one group of male \employers select fewer female \candidates for $k=4,3,2$ choices, namely those younger than 35 years with an annual household income below \$60k and no academic degree ($p_{k=4}=.001 , p_{k=3}=.002, p_{k=2}=.004$). Interestingly, female participants with the same demographic characteristics select significantly more female \candidates ($p_{k=4}<.001 , p_{k=3}<.001, p_{k=2}<.001$). On the other hand, female participants younger than 35 with an income less than \$60k and an academic degree select fewer female \candidates ($p_{k=4}=.008 , p_{k=3}=.005, p_{k=2}=.002$) while male \employers with the same demographic characteristics select more female candidates ($p_{k=4}=.005 , p_{k=2}=.009, p_{k=2}=.007$).
\begin{table}[h]
\centering
\captionsetup{justification=centering}
\resizebox{\columnwidth}{!}{%
\begin{tabular}{ccccc}
  \multicolumn{5}{c}{\textbf{\begin{tabular}[c]{@{}c@{}}Fraction of female candidates per selection\\ (w/ Interactions)\end{tabular}}} \\ \\
\multicolumn{1}{l}{} &
  \multicolumn{1}{l|}{4 Selections} &
  \multicolumn{1}{l|}{3 Selection} &
  \multicolumn{1}{l|}{2 Selections} &
  \multicolumn{1}{l}{1 Selection} \\ 
  \multicolumn{1}{l}{} &
  \multicolumn{1}{c|}{$(k=4)$} &
  \multicolumn{1}{c|}{$(k=3)$} &
  \multicolumn{1}{c|}{$(k=2)$} &
  \multicolumn{1}{c}{$(k=1)$} \\ \hline
(Intercept) &
  \multicolumn{1}{c|}{\textbf{0.454***}} &
  \multicolumn{1}{c|}{\textbf{0.457***}} &
  \multicolumn{1}{c|}{\textbf{0.444***}} &
  \textbf{0.463***} \\
Income $<$\$60k &
  \multicolumn{1}{c|}{\textbf{0.146*}} &
  \multicolumn{1}{c|}{0.141} &
  \multicolumn{1}{c|}{\textbf{0.179*}} &
  0.184 \\
Age $<$ 35 years + Income $<$\$60k &
  \multicolumn{1}{c|}{\textbf{-0.284**}} &
  \multicolumn{1}{c|}{\textbf{-0.279**}} &
  \multicolumn{1}{c|}{\textbf{-0.301*}} &
  -0.252 \\
Age $<$ 35 years + Income $<$\$60k + isAcademic &
  \multicolumn{1}{c|}{\textbf{0.279*}} &
  \multicolumn{1}{c|}{\textbf{0.265*}} &
  \multicolumn{1}{c|}{\textbf{0.319*}} &
  0.330 \\ \hline
isFemale + Age $<$ 35 years &
  \multicolumn{1}{c|}{-0.255} &
  \multicolumn{1}{c|}{\textbf{-0.287*}} &
  \multicolumn{1}{c|}{\textbf{-0.333*}} &
  -0.176 \\
isFemale + Income $<$\$60k &
  \multicolumn{1}{c|}{-0.180} &
  \multicolumn{1}{c|}{-0.175} &
  \multicolumn{1}{c|}{\textbf{-0.239*}} &
  -0.214 \\
isFemale + Age $<$ 35 years + Income $<$\$60k &
  \multicolumn{1}{c|}{\textbf{0.457**}} &
  \multicolumn{1}{c|}{\textbf{0.505***}} &
  \multicolumn{1}{c|}{\textbf{0.620***}} &
  0.473 \\
isFemale + Age $<$ 35 years + Income $<$\$60k + isAcademic &
  \multicolumn{1}{c|}{\textbf{-0.395*}} &
  \multicolumn{1}{c|}{\textbf{-0.420*}} &
  \multicolumn{1}{c|}{\textbf{-0.553*}} &
  -0.520 \\ \hline
   &
  \multicolumn{1}{l}{} &
  \multicolumn{1}{l}{} &
  
\end{tabular}%
}
\caption{Coefficients of a linear regression
model which predicts the fraction of selected female \candidates of the top $k$ \candidates. Input features are \employer features (See Figure \ref{fig:demographics}). This model captures three-way feature interactions. For readability, we only show interactions that are statistically significant (in bold). Note that the Intercept represents baseline regression coefficient for a male \employers older than 35 years, with an income greater than \$60k and a high school degree or no degree. \textit{$*** = p<\frac{0.001}{4} ; **= p<\frac{0.01}{4} ; *=\frac{0.05}{4}$}.
}
\label{tbl:bias-demographics}
\end{table}
\subsection{Employer Understanding of Fairness} \label{sec:employers_fairness}
The textual responses of a few of our participants (proxy \employers) revealed that they are consciously trying to enforce demographic parity when making hiring decisions. 
To confirm this hypothesis, we took a random sample of 100 free form text responses submitted by our participants. We ensured that the responses we sampled are substantive in that each response comprised of at least 200 characters. We then manually inspected these responses and computed what fraction of these participants indicated that they explicitly tried to enforce gender parity when making decisions. 
Our analysis revealed that $12.4\%$ of participants (proxy \employers) actively tried to enforce demographic parity when making decisions and stated so explicitly in the text field of our survey. 

Our analysis also revealed that some of the participants knowingly applied their own notions of fairness not only within tasks (e.g. selecting from the bottom to ``give inexperienced candidates a chance") but also across tasks by discriminating against a group of \candidates in one \jobcontext and ``making up for it" in another \jobcontext. 
Furthermore, 35\% of our respondents explicitly expressed preference for \emph{individual fairness} i.e., treating similar individuals similarly without consideration for gender~\cite{dwork2012fairness}, in their text responses. 
Recall that we also asked each participant to rate the importance of gender (and other displayed features) for each \jobcontext on a 5-point Likert scale. Participant responses revealed that  majority of respondents (59\%, 67\%, and 66\% in case of moving assistance, event staffing, and shopping, respectively) report that they did not consider gender at all when making decisions. These findings suggest that it would be important to investigate how employers' mental models of fairness interact with fair ranking algorithms and impact hiring decisions in the real world. 


\section{Discussion \& Conclusion}\label{sec:conclusions}
In this work, we study how gender biases percolate in online hiring platforms and how they impact real world hiring decisions. More specifically, we analyze how various sources of gender biases in online hiring platforms such as the \jobcontext, 
\candidate profiles, and inherent biases of \employers interact with each other and with ranking algorithms to affect hiring decisions. 

Our analysis revealed 
that fair ranking algorithms can be helpful in increasing the number of \underrepresented \candidates selected. 
However, their effectiveness is dampened in those \jobcontexts where \employers have a persistent gender preference (e.g., moving assistance jobs). Our results also revealed that fair ranking is more effective when \underrepresented \candidate profiles (features) are similar to those in the overrepresented group. Analyzing the textual responses of the study participants also revealed that several of the participants were cognizant of (un)fairness and possible underrepresentation. Furthermore, some of the participants seem to be actively enforcing their preferred notions of fairness when making decisions. 
This work paves way for several interesting future research directions. Firstly, this work underscores the importance of investigating how employers' mental models of fairness interact with ranking algorithms and impact hiring decisions in online and real world settings. Secondly, while this work explicitly focuses on gender biases, it would also be interesting to systematically study the effect of racial biases in online hiring scenarios. 

\bibliography{main}
\newpage
\appendix
\section{Appendix}\label{appendix}
\subsection{Male Candidates as Underrepresented Group}
Table \ref{tbl:algo-compare-selection-rate f>m} shows the selection rates for all algorithms and sets with swapped gender. These experiments serve as the counterfactual scenario to table \ref{tbl:algo-compare-selection-rate m>f}. In this setting, the rankings consist of 7 female candidates and 3 male candidates.
Table \ref{tbl:algo-compare-selection-rate f>m} shows that \fairswapped increases the selection rates for underrepresented male candidates in all \jobcontexts.
The highest selection rates being in the job context moving with an average of 30.90\% over all sets. This is the opposite to the real scenario where women are underrepresented and achieve the lowest selection rates for the moving assistance job. However, underrepresented men achieve overall lower selection rates for the tasks event staffing and shopping. As for female candidates, the randomly ordered ranking and the fair ranking do not achieve better results over 4 selections for the set D2 in the event staffing \jobcontext.
\begin{table}[h!]
\centering
\resizebox{\columnwidth}{!}{%
\begin{tabular}{|c|c|ccc}
\hline
\multicolumn{5}{|c|}{3 male candidates 7 female candidates}                                                \\ \hline
 &
  \textbf{\begin{tabular}[c]{@{}c@{}}Mean selected male\\ candidates per user\end{tabular}} &
  \multicolumn{1}{c|}{Moving assistance} &
  \multicolumn{1}{c|}{Event staffing} &
  \multicolumn{1}{c|}{Shopping} \\ \hline
Moving assistance & 1.16 & \multicolumn{1}{c|}{-} & \multicolumn{1}{c|}{0.10*} & \multicolumn{1}{c|}{0.04} \\ \cline{1-2} \cline{4-5} 
Event staffing    & 1.06 & -                      & \multicolumn{1}{c|}{-}     & \multicolumn{1}{c|}{0.05} \\ \cline{1-2} \cline{5-5} 
Shopping          & 1.11 & -                      & -                          & -                         \\ \cline{1-2}
\end{tabular}%
}
\caption{Average selected male candidates per user compared between job contexts. For the scenarios with swapped genders $***=p<0.01 ; **=p<0.05 ; *= p<0.1 ; .=p<0.15$}
    \label{tbl:context-comparison f>m}
\end{table}
\begin{table}[h!]
\centering
\resizebox{\columnwidth}{!}{%
\begin{tabular}{|c|c|ccc}
\hline
\multicolumn{5}{|c|}{3 female candidates 7 male candidates}                                                     \\ \hline
 &
  \textbf{\begin{tabular}[c]{@{}c@{}}Mean selected female\\ candidates per user\end{tabular}} &
  \multicolumn{1}{c|}{Moving assistance} &
  \multicolumn{1}{c|}{Event staffing} &
  \multicolumn{1}{c|}{Shopping} \\ \hline
Moving assistance & 1.06 & \multicolumn{1}{c|}{-} & \multicolumn{1}{c|}{0.18***} & \multicolumn{1}{c|}{0.17***} \\ \cline{1-2} \cline{4-5} 
Event staffing    & 1.24 & -                      & \multicolumn{1}{c|}{-}       & \multicolumn{1}{c|}{0.01}    \\ \cline{1-2} \cline{5-5} 
Shopping          & 1.23 & -                      & -                            & -                            \\ \cline{1-2}
\end{tabular}%
}
\caption{Average selected female candidates per user compared between job contexts. $***=p<0.01 ; **=p<0.05 ; *= p<0.1 ; .=p<0.15$}
    \label{tbl:context-comparison m>f}
\end{table}
\subsection{Selected Underrepresented Candidates per User}
Table \ref{tbl:context-comparison m>f} and table \ref{tbl:context-comparison f>m} show the average number of selected underrepresented \candidates and compare the differences between the job contexts.\\
In the traditional setting with female candidates as underrepresented group, table \ref{tbl:context-comparison m>f} shows the significant differences between moving assistance compared to event staffing and shopping. Out of 4 selections, users chose on average 1.06 women for the task moving assistance, 1.24 women for the task event staffing and 1.23 women for the task shopping.\\
In the counterfactual setting with male \candidates as underrepresented group, we can see that the selection rates are generally lower. The only exception being the \jobcontext moving assistance with 1.16 selected male candidates per user on average.
\subsection{\fair}
Table \ref{tbl:ranking-fair-ordering} shows the fair ordering for the retrieved sets (D1, D2, D3) in both settings. All underrepresented candidates achieve better rankings through the fair ordering.
\begin{table}
\centering
\resizebox{\columnwidth}{!}{%
\begin{tabular}{|l|l|l|l|l|l|l|l|l|l|l|}
\hline
\textbf{Rank}                   & \textbf{1} & \textbf{2} & \textbf{3} & \textbf{4} & \textbf{5} & \textbf{6} & \textbf{7} & \textbf{8} & \textbf{9} & \textbf{10} \\ \hline
\textbf{\fair(0.42, 0.58)} & m          & f          & m          & f          & m          & f          & m          & m          & m          & m           \\ \hline
\textbf{\fairswapped(0.58, 0.42)} & f & m & f & m & f & m & f & f & f & f\\ \hline
\end{tabular}%
}
\vspace{5pt}
\caption{Fair ordering for all data sets using \fair with $p_{male}=0.58$ and $p_{female}=0.42$. and \fairswapped with $p_{male}=0.42$ and $p_{female}=0.58$}
\label{tbl:ranking-fair-ordering}
\end{table}
\begin{table*}[t!]
\resizebox{\textwidth}{!}{%
\begin{tabular}{|c|c|c|c|c|c|c|c|c|c|c|c|c|}
\hline
\multicolumn{13}{|c|}{\cellcolor{ba}\textbf{Moving assistance}}                                                                                                                     \\ \hline
\cellcolor{ba}\textbf{}          & \multicolumn{4}{c|}{\cellcolor{ba}\textbf{Task Rabbit\_swapped}} & \multicolumn{4}{c|}{\cellcolor{ba}\textbf{Random\_swapped}} & \multicolumn{4}{c|}{\cellcolor{ba}\textbf{Fair\_swapped}} \\ \hline
\cellcolor{ba}\textbf{\#Choices} & \cellcolor{ba}\textbf{1}           & \cellcolor{ba}\textbf{2}          & \cellcolor{ba}\textbf{3}          &\cellcolor{ba}\textbf{ 4}          &\cellcolor{ba}\textbf{ 1}         &\cellcolor{ba}\textbf{ 2 }        & \cellcolor{ba}\textbf{3 }        &\cellcolor{ba}\textbf{ 4}         & \cellcolor{ba}\textbf{1 }        &\cellcolor{ba}\textbf{ 2}         &\cellcolor{ba}\textbf{ 3 }       & \cellcolor{ba}\textbf{4}        \\ \hline
\cellcolor{ba}\textbf{D1}        & 5.08\%      & 9.32\%     & 14.69\%    & 15.25\%    & \cellcolor{5-10}14.75\%   &\cellcolor{5-10} 16.39\%   &\cellcolor{<5} 15.85\%   &\cellcolor{<5} 16.39\%   &\cellcolor{5-10} 10.53\%   &\cellcolor{5-10} 18.42\%   &\cellcolor{5-10} 24.56\%  &\cellcolor{5-10} 23.25\%  \\ \hline
\cellcolor{ba}\textbf{D2}        & 38.33\%     & 48.33\%    & 52.22\%    & 45.00\%    & \cellcolor{10-15}48.39\%   &\cellcolor{<5} 51.61\%   &\cellcolor{<5} 54.84\%   &\cellcolor{<5} 48.79\%   &\cellcolor{5-10} 48.28\%   &\cellcolor{<5} 51.72\%   &\cellcolor{5-10} 57.47\%  &\cellcolor{<5} 48.28\%  \\ \hline
\cellcolor{ba}\textbf{D3}        & 8.33\%      & 10.00\%    & 14.44\%    & 18.33\%    & \cellcolor{5-10}15.00\%   &\cellcolor{5-10} 16.67\%   &\cellcolor{5-10} 21.67\%   &\cellcolor{5-10} 23.75\%   &\cellcolor{10-15} 18.64\%   &\cellcolor{5-10} 15.25\%   &\cellcolor{<5} 16.38\%  &\cellcolor{<5} 21.19\%  \\ \specialrule{.1em}{.1em}{.1em}
\cellcolor{ba}\textbf{all}       & 17.25\%     & 22.55\%    & 27.12\%    & 26.20\%    & \cellcolor{<5}26.05\%   &\cellcolor{5-10} 28.22\%   &\cellcolor{<5} 30.78\%   &\cellcolor{<5} 29.64\%   &\cellcolor{5-10} 25.82\%   &\cellcolor{5-10} 28.47\%   &\cellcolor{5-10} 32.81\%  &\cellcolor{<5} 30.90\%  \\ \specialrule{.1em}{.1em}{.1em}
\multicolumn{13}{|c|}{\cellcolor{ba}\textbf{Event Staffing}}                                                                                                                        \\ \hline
\textbf{\cellcolor{ba}}          & \multicolumn{4}{c|}{\cellcolor{ba}\textbf{Task Rabbit\_swapped}} & \multicolumn{4}{c|}{\cellcolor{ba}\textbf{Random\_swapped}} & \multicolumn{4}{c|}{\cellcolor{ba}\textbf{Fair\_swapped}} \\ \hline
\cellcolor{ba}\textbf{\#Choices} &\cellcolor{ba}\textbf{ 1}           &\cellcolor{ba}\textbf{ 2}          &\cellcolor{ba}\textbf{ 3 }         &\cellcolor{ba}\textbf{ 4}          &\cellcolor{ba}\textbf{ 1}         &\cellcolor{ba}\textbf{ 2 }        &\cellcolor{ba}\textbf{ 3}         & \cellcolor{ba}\textbf{4}         & \cellcolor{ba}\textbf{1}         &\cellcolor{ba}\textbf{ 2}         & \cellcolor{ba}\textbf{3}        &\cellcolor{ba}\textbf{ 4 }       \\ \hline
\cellcolor{ba}\textbf{D1}        & 6.67\%      & 12.50\%    & 14.44\%    & 15.84\%    & \cellcolor{5-10}13.33\%   &\cellcolor{<5} 15.00\%   &\cellcolor{<5} 16.67\%   &\cellcolor{<5} 17.50\%   &\cellcolor{<5} 8.47\%    &\cellcolor{<5} 16.95\%   &\cellcolor{<5} 17.51\%  &\cellcolor{<5} 17.80\%  \\ \hline
\cellcolor{ba}\textbf{D2}        & 37.29\%     & 46.61\%    & 55.37\%    & 45.76\%    & \cellcolor{5-10}42.62\%   &\cellcolor{<5} 45.08\%   &\cellcolor{-5-10} 50.27\%   &\cellcolor{-5} 43.85\%   &\cellcolor{>15} 52.63\%   &\cellcolor{-5} 45.61\%   &\cellcolor{<5} 55.56\%  &\cellcolor{-5} 44.74\%  \\ \hline
\cellcolor{ba}\textbf{D3}        & 1.67\%      & 1.67\%     & 6.67\%     & 14.58\%    & \cellcolor{<5}6.45\%    &\cellcolor{5-10} 12.90\%   &\cellcolor{5-10} 17.74\%   &\cellcolor{<5} 19.35\%   &\cellcolor{<5} 5.17\%    &\cellcolor{5-10} 7.76\%    &\cellcolor{5-10} 16.09\%  &\cellcolor{5-10} 20.26\%  \\ \specialrule{.1em}{.1em}{.1em}
\cellcolor{ba}\textbf{all}       & 15.21\%     & 20.26\%    & 25.49\%    & 25.39\%    & \cellcolor{5-10}20.80\%   &\cellcolor{<5} 24.33\%   &\cellcolor{<5} 28.23\%   &\cellcolor{<5} 26.90\%   &\cellcolor{5-10} 22.09\%   &\cellcolor{<5} 23.44\%   &\cellcolor{<5} 29.72\%  &\cellcolor{<5} 27.60\% \\ \specialrule{.1em}{.1em}{.1em}
\multicolumn{13}{|c|}{\cellcolor{ba}\textbf{Shopping}}                                                                                                                              \\ \hline
\cellcolor{ba}\textbf{}          & \multicolumn{4}{c|}{\cellcolor{ba}\textbf{Task Rabbit\_swapped}} & \multicolumn{4}{c|}{\cellcolor{ba}\textbf{Random\_swapped}} & \multicolumn{4}{c|}{\cellcolor{ba}\textbf{Fair\_swapped}} \\ \hline
\cellcolor{ba}\textbf{\#Choices} & \cellcolor{ba}\textbf{1}           &\cellcolor{ba}\textbf{ 2}          &\cellcolor{ba}\textbf{ 3}          &\cellcolor{ba}\textbf{ 4 }         &\cellcolor{ba}\textbf{ 1}         &\cellcolor{ba}\textbf{ 2}         & \cellcolor{ba}\textbf{3}         &\cellcolor{ba}\textbf{ 4 }        & \cellcolor{ba}\textbf{1}         &\cellcolor{ba}\textbf{ 2}         &\cellcolor{ba}\textbf{ 3}        &\cellcolor{ba}\textbf{ 4}        \\ \hline
\cellcolor{ba}\textbf{D1}        & 5.00\%      & 6.67\%     & 13.89\%    & 17.08\%    & \cellcolor{-5}4.84\%    &\cellcolor{5-10} 13.71\%   &\cellcolor{<5} 18.82\%   &\cellcolor{<5} 20.16\%   &\cellcolor{5-10} 13.79\%   &\cellcolor{5-10} 18.10\%   &\cellcolor{<5} 18.39\%  &\cellcolor{<5} 18.53\%  \\ \hline
\cellcolor{ba}\textbf{D2}        & 38.33\%     & 45.83\%    & 49.44\%    & 45.00\%    & \cellcolor{>15}53.33\%   &\cellcolor{5-10} 53.33\%   &\cellcolor{10-15} 60.00\%   &\cellcolor{<5} 49.58\%   &\cellcolor{10-15} 49.15\%   &\cellcolor{10-15} 50.85\%   &\cellcolor{5-10} 54.80\%  &\cellcolor{<5} 47.88\%  \\ \hline
\cellcolor{ba}\textbf{D3}        & 5.08\%      & 7.63\%     & 10.17\%    & 14.83\%    & \cellcolor{5-10}14.75\%   &\cellcolor{<5} 12.30\%   &\cellcolor{<5} 12.02\%   &\cellcolor{<5} 15.98\%   &\cellcolor{5-10} 14.04\%   &\cellcolor{5-10} 16.67\%   &\cellcolor{5-10} 17.54\%  &\cellcolor{<5} 19.74\%  \\ \specialrule{.1em}{.1em}{.1em}
\cellcolor{ba}\textbf{all}       & 16.14\%     & 20.04\%    & 24.50\%    & 25.64\%    & \cellcolor{5-10}24.31\%   &\cellcolor{5-10} 26.45\%   &\cellcolor{5-10} 30.28\%   &\cellcolor{<5} 28.58\%   &\cellcolor{5-10} 25.66\%   &\cellcolor{5-10} 28.54\%   &\cellcolor{5-10} 30.25\%  &\cellcolor{<5} 28.72\%  \\ \hline
\end{tabular}%
}
\caption{Male selection rates for the cumulative selections 1-4. Colors indicate the change in ppt compared to \rabbitswapped:\\
\colorbox{-5-10}{[-10,-5]} \colorbox{-5}{(-5,0)}|0|\colorbox{<5}{(0,+5)}|\colorbox{5-10}{[+5,+10)}|\colorbox{10-15}{[+10,+15)}|\colorbox{>15}{[+15,$\infty$)}}\label{tbl:algo-compare-selection-rate f>m}

\end{table*}

\end{document}